\documentclass[journal]{vgtc}                

\ifpdf
  \pdfoutput=1\relax                   
  \usepackage{graphicx}                
  \DeclareGraphicsExtensions{.pdf,.png,.jpg,.jpeg} 
\else
  \usepackage{graphicx}                
  \DeclareGraphicsExtensions{.eps}     
\fi%

\graphicspath{{figures/}{images/pei/}{images/pei_ppt/}{images/dawn/}{images/dawn/eva1/}{images/dawn/case2/}{images/dawn/case3/}{./}} 

\usepackage{microtype}                 
\PassOptionsToPackage{warn}{textcomp}  
\usepackage{textcomp}                  
\usepackage{mathptmx}                  
\usepackage{amsmath}
\usepackage{times}                     
\usepackage{setspace}
\usepackage[colorlinks,linkcolor=blue,bookmarks=false]{hyperref}
\usepackage{algorithm}
\usepackage{algpseudocode}
\usepackage{tabularx}
\usepackage{tabu,multirow,makecell,booktabs}

\usepackage{cite}                      
\usepackage{tabu}                      
\usepackage{booktabs}                  
\usepackage{enumerate}
\usepackage[caption=false]{subfig}
\usepackage{etoolbox,lipsum}
\usepackage{setspace}
\usepackage[left=1.2cm,top=1.7cm,bottom=1.6cm,right=1.2cm]{geometry}

\onlineid{1157}

\vgtccategory{Research}
\vgtcpapertype{algorithm/technique}

\title{VisCode: Embedding Information in Visualization Images using Encoder-Decoder Network}

\author{Peiying Zhang, Chenhui Li, and Changbo Wang}
\authorfooter{
\item
Peiying Zhang, Chenhui Li, and Changbo Wang are with School of Computer Science and Technology, East China Normal University. Chenhui Li is the corresponding author. E-mail: chli@cs.ecnu.edu.cn.
}

\shortauthortitle{Embedding Information in Visualization Images using Encoder-Decoder Network}

\abstract{We present an approach called VisCode for embedding information into visualization images. This technology can implicitly embed data information specified by the user into a visualization while ensuring that the encoded visualization image is not distorted. The VisCode framework is based on a deep neural network. We propose to use visualization images and QR codes data as training data and design a robust deep encoder-decoder network. The designed model considers the salient features of visualization images to reduce the explicit visual loss caused by encoding. To further support large-scale encoding and decoding, we consider the characteristics of information visualization and propose a saliency-based QR code layout algorithm. We present a variety of practical applications of VisCode in the context of information visualization and conduct a comprehensive evaluation of the perceptual quality of encoding, decoding success rate, anti-attack capability, time performance, etc. The evaluation results demonstrate the effectiveness of VisCode.
} 


\teaser{
  \centering
  \includegraphics[width=1.0\linewidth]{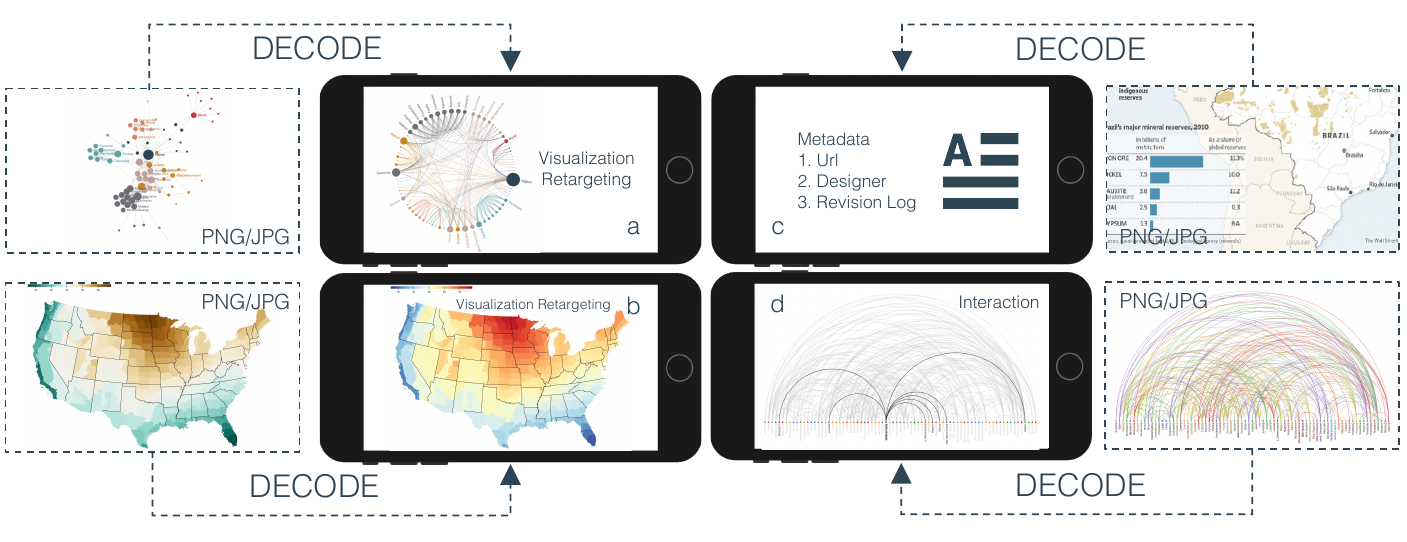}
  \vspace{-6pt}
  \caption{VisCode can be used in a variety of practical applications. The input is a static visualization image. According to the encoding rules, we can retarget the original visual style after performing the decoding operation. Visualization retargeting includes representation (a) and theme color (b). VisCode can be also applied to metadata embedding as shown in (c) and large-scale information embedding. After decoding the source code into a static image, we can quickly build a visualization application and apply a common interaction such as selection as shown in (d).
  }
  \label{fig:pipeline}
}

\vgtcinsertpkg

\begin{document}
\firstsection{Introduction}
\maketitle

For a long time, information visualizations were created by artists, designers or programmers, and most of them were disseminated in the form of bitmap images. Many researchers have performed many studies on specific aspects of the design of information representations, such as the visualization layout, the color theme, and more specific explorations. However, with the development of fast network technology, the spread of visualization images has encountered two problems. One is that the copyright problem for visualization has not been effectively solved. Most designers use explicit icons to present copyright information, but this method affects the aesthetics of a chart and can obstruct important information. Moreover, when a visualization is converted into a bitmap image, the data in the visualization are converted into pixels, and the original data are completely lost. Although several researchers, such as Poco et al.~\cite{poco2017extracting}, have applied a pattern recognition approach to visualization images to retarget the color theme, the effectiveness of data recognition is still related to the form of visualization. As an alternative approach, information steganography~\cite{cox2007digital} (a means of implicit information embedding) for visualization images has potential for copyright protection and source data preservation.

Information steganography has a very meaningful practical application in the field of visualization design.
In many forms of information propagation, such as social networks, blogs, and magazines, bitmap images are used most frequently and do not require the support of an additional web server. After conversion into an image, the original code used to generate the image is lost, thus increasing the difficulty of modification and visualization redesign. Similar difficulties also arise when designers and programmers collaborate. However, if a designer could implicitly embed the key visualization code into an image during the visualization design period, then the dissemination and revision of the visualized information would not require two different types of media (image and text), which would help programmers integrate such visualizations into big data analysis systems. Another potential application is the visualization retargeting. Visualization retargeting is very useful for the creation of visualizations. On the one hand, it reduces the designer's workload. On the other hand, it broadens the artist's creative space. \autoref{fig:pipeline} shows four practical applications of our work.

Information steganography is different from digital watermarking, which is a technique for hiding information in a piece of media (usually an image). As described in the book of Cox et al.~\cite{cox2007digital}, steganography has a long history. Researchers have proposed a variety of methods of embedding information in images for diverse steganographic applications. As surveyed by Ghadirli et al.~\cite{ghadirli2019overview}, the techniques used in image steganography include chaos-based methods, permutation-based methods, frequency-domain methods, hash-based methods, and evolutionary methods, among others. The continuous development of steganographic technology and the recent emergence of deep learning technology have enabled the extension of image steganography into broader areas of application. To our knowledge, however, few researchers have studied how to embed information in visualization images. Related research has focused only on the recognition of visualization images~\cite{poco2017reverse,haehn2018evaluating}. Although many methods of steganography have been proposed for natural images, the features of data visualizations are quite distinct. First, most of these models are based on natural images and make use of rich and mature features, while visualization charts usually have clean backgrounds and clear visual elements, increasing the difficulty of encoding and decoding. Second, Bylinskii et al.~\cite{bylinskii2017learning} demonstrated that the visual importance of visualizations depends not only on the image context but also on higher-level factors (e.g., semantic categories), which is different from that of natural images.


We present a novel information embedding approach for visualization images, called VisCode. VisCode is an end-to-end autoencoding method that implicitly encodes information into a visualization image with little visual error. First, we propose to use QR codes as a form of media in which to store information to increase the success rate of decoding. We create a dataset consisting of visualization chart images and QR codes to be used for autoencoder training. Second, we outline a deep neural network architecture based on saliency detection that is designed to perform stable information steganography for visualization images. Information in the form of QR codes is adaptively encoded into the visually nonsignificant areas of a visualization image by means of a saliency-based layout algorithm to reduce the visual traces of this encoding. We also report the successful application of the VisCode method to three practical application scenarios: metadata embedding, source code embedding, and visualization retargeting.
Based on the embedded information, the designer can easily manage and modify a graphical chart produced using this method, for example, to generate different versions of the visual design. Additionally, our VisCode system can convert static visualization images into interactive images to facilitate more accurate visualization comprehension. Furthermore, our tool supports visualization retargeting and improvement in accordance with user preferences, such as modification of the color palette style and the form of visualization.

Our experimental results and evaluations show that VisCode has great potential for application in information visualization. In summary, our contributions include three aspects:
\begin{itemize}
\setlength{\itemsep}{0pt}
\setlength{\parsep}{0pt}
\setlength{\parskip}{0pt}
\item [(1)] We define the problem of information embedding in the context of information visualization. We verify the practical importance of this study in pioneering a new application domain.
\item [(2)] We outline a deep learning framework that can be used to implement personalized steganography for information visualization images. This framework can achieve high-quality large-scale information hiding.
\item [(3)] We present a set of evaluations to demonstrate the effectiveness of our proposed framework from multiple perspectives, including user friendliness,
indices of encoding quality and decoding success, steganography defense, and time performance.
\end{itemize}


\section{Related Work}
\label{sec:rel}

\subsection{Information Steganography}
Information steganography, the art of concealing secret data within other carrier data, has numerous applications (e.g., digital watermarking, covert transmission, and access control). The most popular form of steganography is to embed secret information by slightly perturbing carrier images, videos, 3D models, texts, or audio data while ensuring that these perturbations are imperceptible to the human visual system.
Instead of processing popular forms of media such as videos and images, COBRA~\cite{hao2012cobra} encodes data into a special type of colorful 2D barcode to improve the efficiency of transmission between a smartphone screen and camera. Yang et al.~\cite{yang20163d} outlined an effective steganography algorithm using a 3D geometric model as the carrier. Their algorithm hides information in the histogram of the vertex coordinates of the 3D model and can achieve robust performance.
Jo et al.~\cite{jo2016disco} utilized light signals from displays and cameras to transmit information.
Xiao et al.~\cite{xiao2018fontcode} proposed an interesting method of directly embedding messages into the letters of text. The glyphs of fonts are perturbed to encode secret information, and this information is decoded by recognizing characters by means of an optical character recognition (OCR) library. However, this method has two limitations: different perturbation rules are needed when processing different font types, and sufficient resolution of the text document is necessary when decoding a message.
Hota and Huang~\cite{HotaEmbedding2020} adopt an image watermarking technique to embed the information into a scientific visualization. It seems that their work is similar to VisCode. However, the variety of data types, representations, and propagations of the information visualizations pose new challenges. The steganography work of information visualization is different from scientific visualization.

In summary, previous research on information steganography has investigated a variety of carriers, but there is less previous study on information steganography for information visualization. In the information visualization field, steganographic applications of personalized information visualization require a specially designed information steganography approach that is different from previous methods.

\begin{figure*}[htb]
  \centering
  \includegraphics[width=1.0\linewidth]{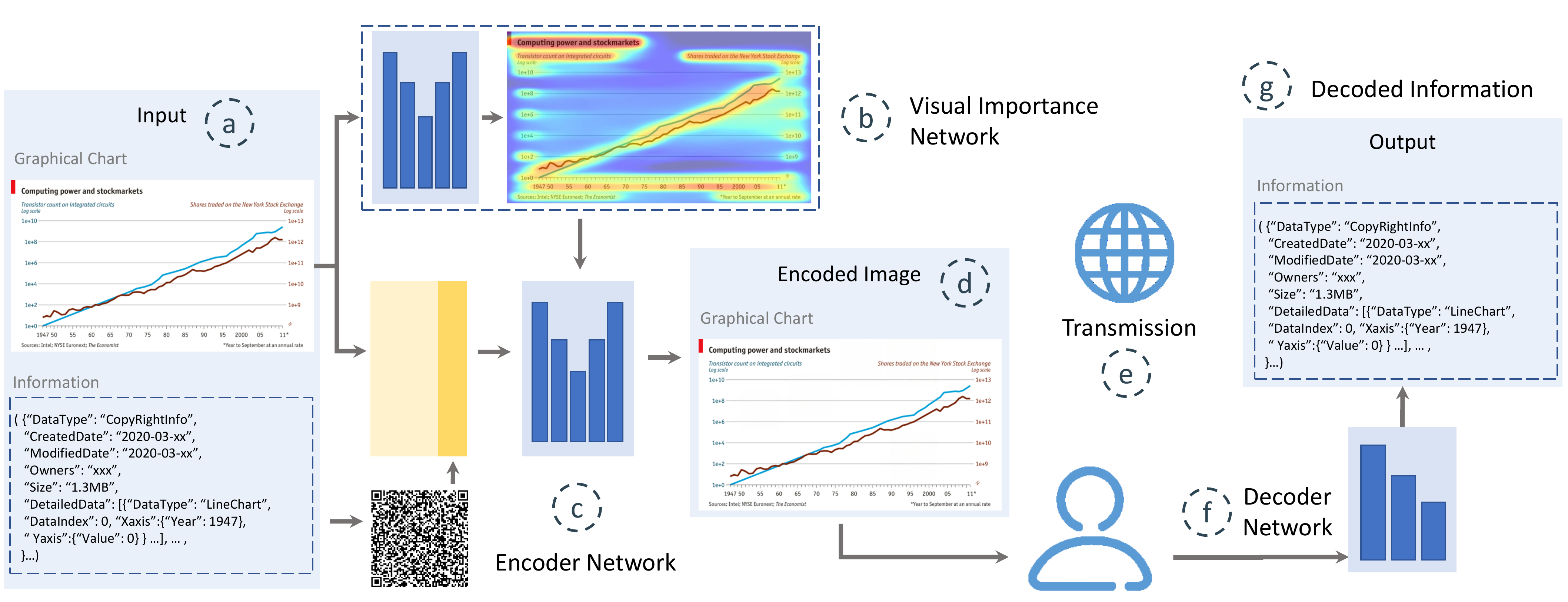}
  \vspace{-12pt}
  \caption{
  The main components of our VisCode system. First, the visual importance network (b) processes the input graphical chart (a), which can facilitate salient feature detection of the visualization and output a visual importance map. Next, an encoder network (c) embeds secret information in the graphical chart (a). The carrier chart image and the QR code image are embedded into vectors by the feature extraction process. Then, these two vectors are concatenated (as the yellow rectangle shown) and fed into the auto-encoding stage. The encoded image (d) is then sent to the user (e), and the user can send it to others by digital transmission. When a user wishes to obtain the detailed information hidden in the chart, the encoded image can be uploaded to the decoder network (f). After data recovery and error correction, the user receives the decoded information (g).
  }
  \label{fig:main_pipeline}
\end{figure*}

\subsection{Image Steganography}
\noindent \textbf{Traditional Steganography}
Images are the most widely processed and exchanged information carriers in the steganographic domain.
Early methods used spatial-domain techniques, such as bit plane complexity segmentation (BPCS)~\cite{kawaguchi1999principles} or least-significant-bit (LSB) substitution~\cite{mielikainen2006lsb}. These methods involve applying small variations to pixel bits, which are not visually obvious. However, they are easily detected by statistical analysis due to their fixed principles of operation.
On the basis of digital signal processing theory, some approaches have been developed for embedding messages in the frequency domain. Almohammad et al.~\cite{almohammad2008high} hid information based on the discrete cosine transform (DCT).
To avoid detection through statistical analysis, many steganography algorithms define special functions to optimize the localization of embeddings. Highly undetectable steganography (HUGO)~\cite{pevny2010using} minimizes distortion by calculating a variable weight for each pixel based on high-dimensional image models. The Wavelet Obtained Weights (WOW) algorithm~\cite{holub2012designing} measures the textural complexity of different image regions with directional filters.

\noindent \textbf{Deep Steganography}
Recent studies have achieved impressive results by combining deep learning with steganography.
Pibre et al.~\cite{pibre2016deep} demonstrated that the results obtained from deep neural networks (DNNs) surpass the conventional results based on handcrafted image features.
Several methods utilize a DNN as a component in combination with traditional steganography techniques~\cite{husien2015artificial, tang2017automatic}. More recently, end-to-end steganography networks have become popular. Inspired by autoencoders, Baluja et al.~\cite{baluja2017hiding} learned feature representations to embed a secret image into a cover image. Hayes et al.~\cite{hayes2017generating} utilized adversarial training to generate steganography results. Some studies have also considered sources of corruption in digital transmission, such as JPEG compression~\cite{zhu2018hidden} and light field messaging (LFM)~\cite{wengrowski2019light}.
Most of these models are based on natural images and make use of rich and mature features. We focus on encoding a large amount of data into images designed for visualization purposes while preserving their visual perceptual quality.

\subsection{Retargeting of Information Visualization}
Early work on information visualization focused on data representations. Tufte~\cite{tufte2001visual} introduced the basic principles of quantitative chart design, emphasizing that the purpose of chart design is to allow users to quickly obtain rich and accurate information and, consequently, decoration that is not related to the data is unnecessary. Tufte's research illustrates the importance of data and the user's understanding of charts in information visualization. In contrast to Tufte, Wilkinson~\cite{wilkinson2012grammar} offered a method of describing charts at a higher level of abstraction. This abstract method involves categorizing charts based on the visual characteristics of the data and geometric figures to simplify the architecture of the drawing system.
In recent years, scattering, network diagrams, regions, dynamics, metaphors, tools, and other visualization techniques have been proposed, and the capabilities of information visualization are becoming increasingly close to ideal. 
With the development of big data and deep learning technologies, visualization retargeting approaches have been presented to aid in visualization redesign. Poco et al.~\cite{poco2017extracting} outlined an approach for extracting color mappings from static visualization images. They evaluated their technique on a variety of visualization types. Later, an end-to-end system was proposed by Poco and Heer~\cite{poco2017reverse} for extracting information from static chart images. Similar works include iVoLVER~\cite{mendez2016ivolver} and Chart Decoder~\cite{dai2018chart}, and Enamul and Maneesh~\cite{SearchingD32020} have similarly offered a deconstruction tool for extracting data and visual encodings from D3 charts. In addition to common usage, functionalities intended to support visually impaired users have also been considered; for example, Choi et al.~\cite{choi2019visualizing} presented a DNN-based method of extracting the features of a chart to help visually impaired users comprehend charts.

To the best of our knowledge, less previous work has addressed steganography in the context of information visualization.
In this paper, we implement visualization retargeting from a novel perspective using information steganography.

\begin{figure}[tbp]
  \centering
  \includegraphics[width=0.7\columnwidth]{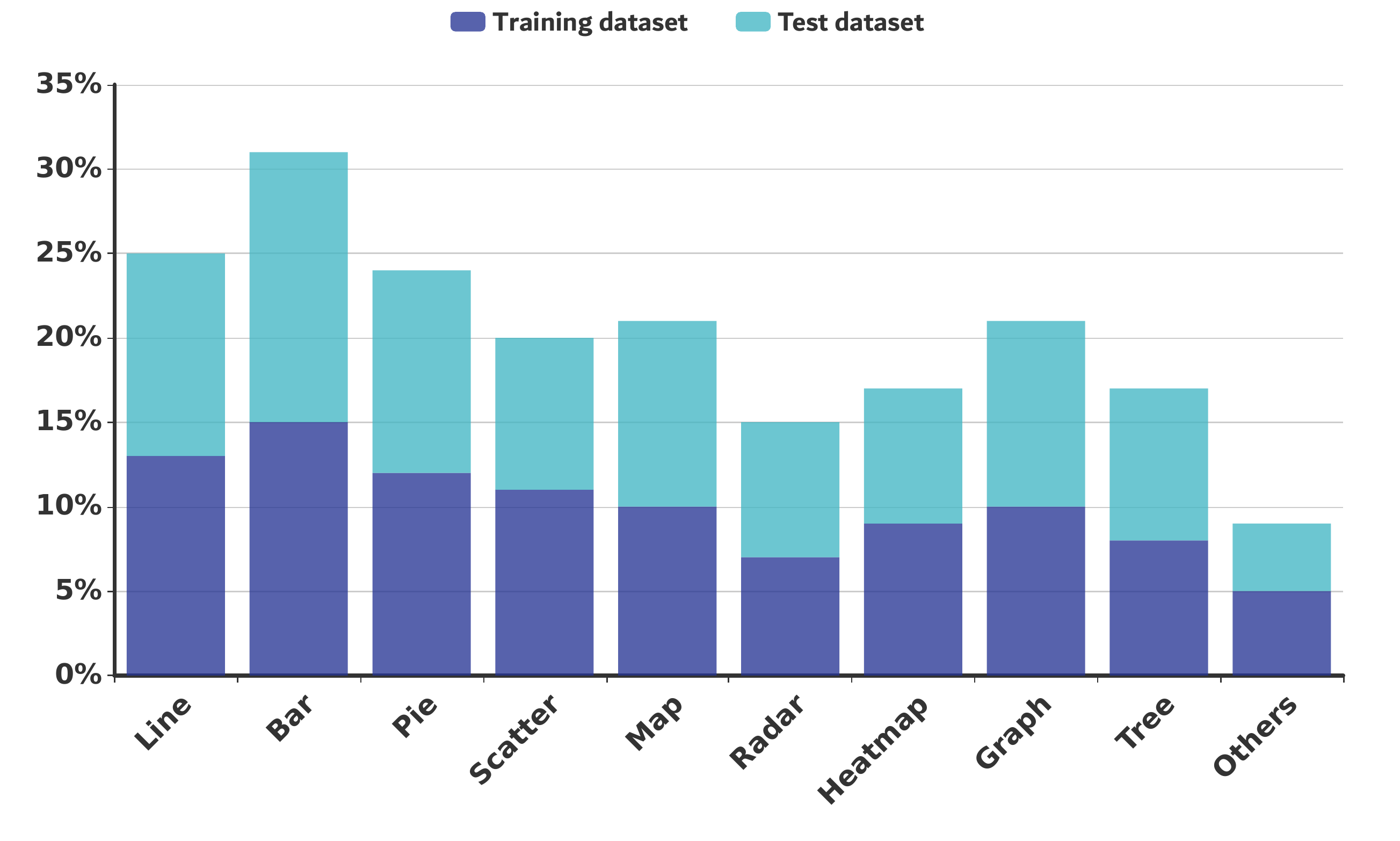}
  \vspace{-10pt}
  \caption{\label{fig:cov_distri}Distribution of different categories in our visualization dataset.}
\end{figure}

\section{Overview}
\label{sec:overview}

An essential goal of data visualization is to improve the cognition and perception of data. Compared with the direct transmission of a large amount of raw data, images are more easily and widely processed carriers for disseminating visual information on the Internet. However, after the original detailed information (e.g., data points, axis text, and color legends) is converted into pixel values, it is difficult to reconstruct the original information from the resulting images. To solve this problem, we attempt to encode additional information into the carrier visual designs with minimal perturbation.

In this paper, we propose VisCode, a framework for embedding and recovering hidden information in a visualization image using an encoder-decoder network. \autoref{fig:main_pipeline} shows the VisCode model, which has 3 main components:
\begin{itemize}
\item \emph{Visual Importance Network} The preparation stage includes a visual importance network and a text transformation model. Since the semantic features of data visualizations are quite different from those of natural images, we use a network to model the visual importance of the input data graph. The resulting importance map is applied as a constraint in the \emph{encoder network}.
Accounting for error correction, we convert the plain message to be embedded into a series of QR codes instead of binary data.

\item \emph{Encoder Network} The encoder network embeds the message within the carrier image and outputs the resulting coded image. In this network component, the carrier chart image and the QR code image are embedded into vectors by the feature extraction process. Then, these two vectors are concatenated and fed into an autoencoder. The resolution of the output coded image is the same as that of the input carrier image.

\item \emph{Decoder Network} The decoder network retrieves the information from the coded image.
Users can share coded images with others through digital transmission. After such a coded image is decoded, the final information is obtained from the decoded QR code image with error correction.

\end{itemize}

\subsection{Definition}
Our VisCode model assumes that, given an information visualization image ${I_c}$ (also called the carrier image) and a plain message ${T_s}$, the coded image ${I_c}^{'}$ output by the encoder network $Enc()$ should be perceptually similar to ${I_c}$ according to human vision, while the recovered message ${T_s}^{'}$ output by the decoder network $Dec()$ should be as accurate as the input ${T_s}$. Considering error correction, we convert the input plain message ${T_s}$ to the QR code image instead of binary data. That is ${I_s} = PrepQR({T_s})$, with $PrepQR()$ being the preparation of QR code from the text, ${I_s}$ being the transformed QR code image (also called the secret image). The output of the decoder network is ${I_s}^{'}$, and ${T_s}^{'} = PostQR({I_s}^{'})$, where $PostQR()$is the process of converting a recovered QR code image into text form with an error-correction coding scheme, and ${T_s}^{'}$ is the result text. Formally, we attempt to learn functions $Enc()$ and $Dec()$ such that:
\begin{equation}
minimize \left\| {{I_c} - {I_c}^{'}} \right\| + \alpha \left\| {{I_s} - {I_s}^{'}} \right\|
\end{equation}
where $Enc({I_c},{I_s}) = {I_c}^{'}$ and $Dec({I_c}^{'}) = {I_s}^{'}$ represent the two functions of interest. $\alpha $ is a scalar weight.

\begin{figure}[tbp]
\centering
\includegraphics[width=0.96\columnwidth]{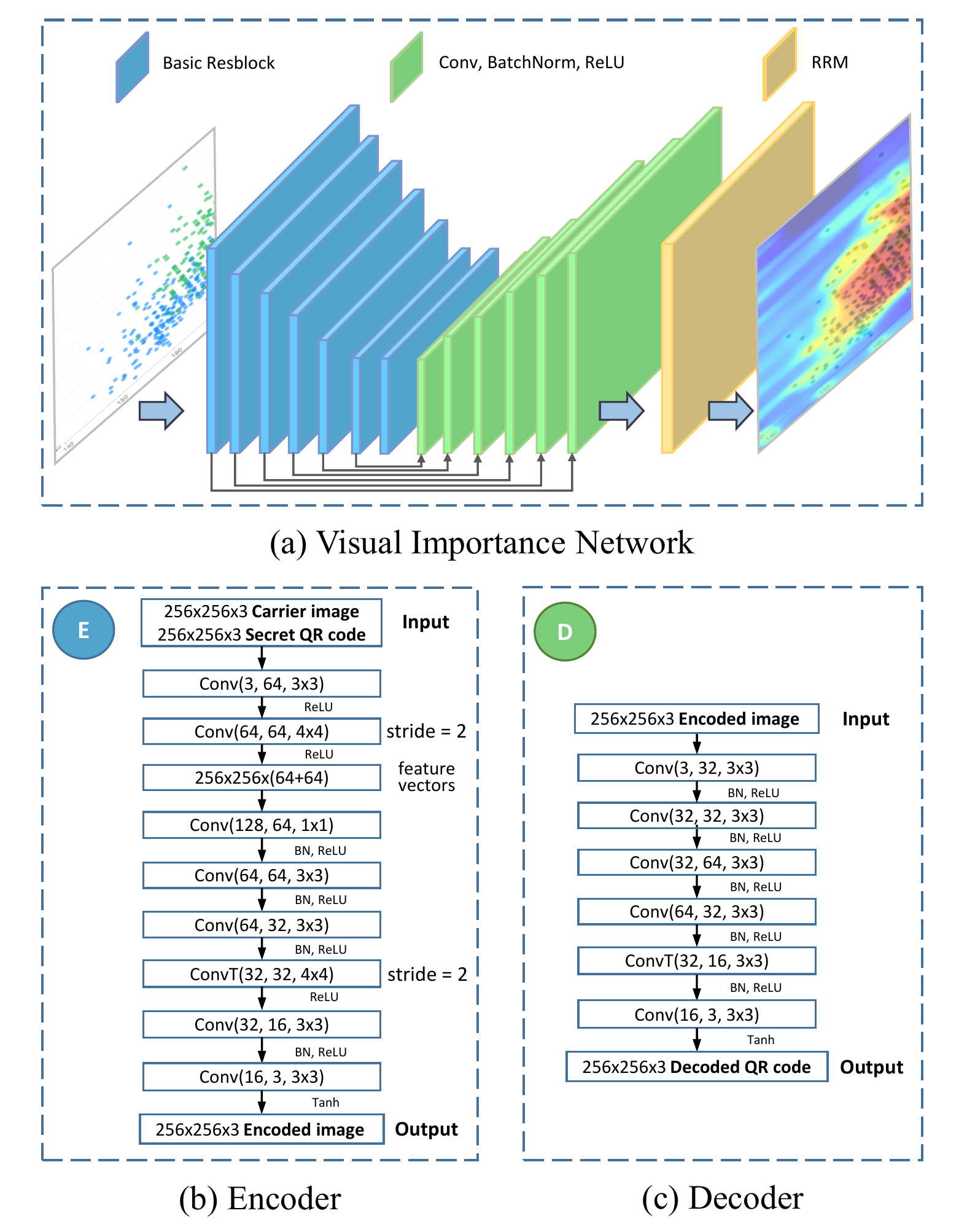}
\vspace{-10pt}
\caption{\label{fig:net_struc}VisCode system contains three network models such as visual importance network, encoder network, and decoder network.}
\end{figure}

\section{Methods}
\label{sec:method}

\subsection{Dataset}
\textbf{Information Visualizations}
\label{subsec:DV_dataset}
The features of data visualization images are quite different from those of natural images. Hence, to build our data visualization dataset for training, we used several popular chart and visualization libraries, such as D3, Echarts, AntV, and Bokeh, to generate carrier visual designs in 10 categories (e.g., ``Bar Chart'' or ``Scatter Chart''). We also selected a subset of images from MASSVIS~\cite{borkin2013makes}, a dataset of visualizations collected from a broad variety of social and scientific domains. The resolutions of these images are expressed as $H \times W$, where $H,W \in [300,3000]$. We selected 1500 images and split them into 1000 images composing the training set and 500 images composing the test set. ~\autoref{fig:cov_distri} shows the distribution of this visualization dataset.

\begin{figure}[tbp]
  \centering
  \subfloat[Original chart]{\includegraphics[width=.225\linewidth]{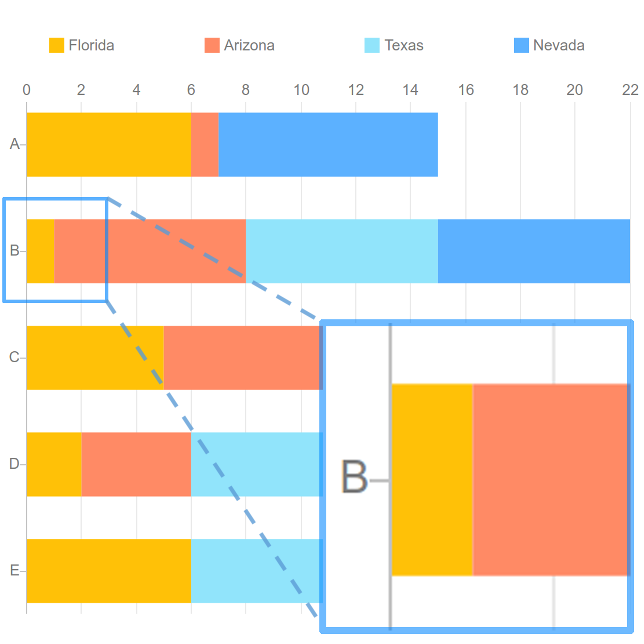}}
  \hfil
  \subfloat[VisCode]{\includegraphics[width=.225\linewidth]{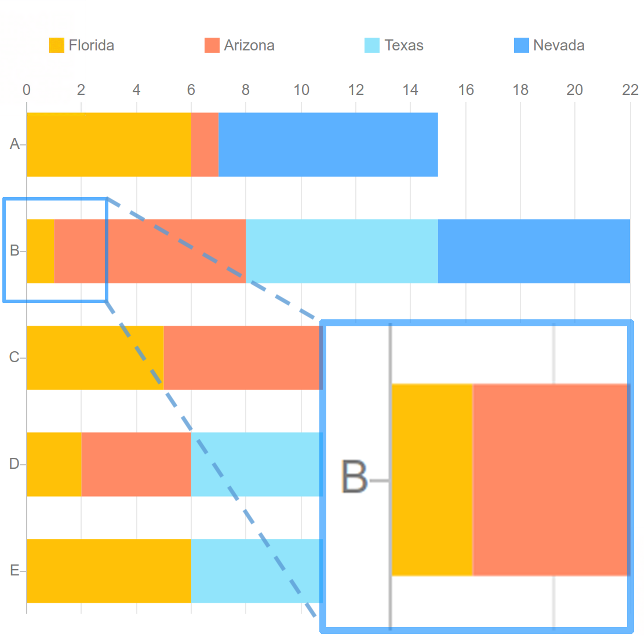}}
  \hfil
  \subfloat[SteganoGAN]{\includegraphics[width=.225\linewidth]{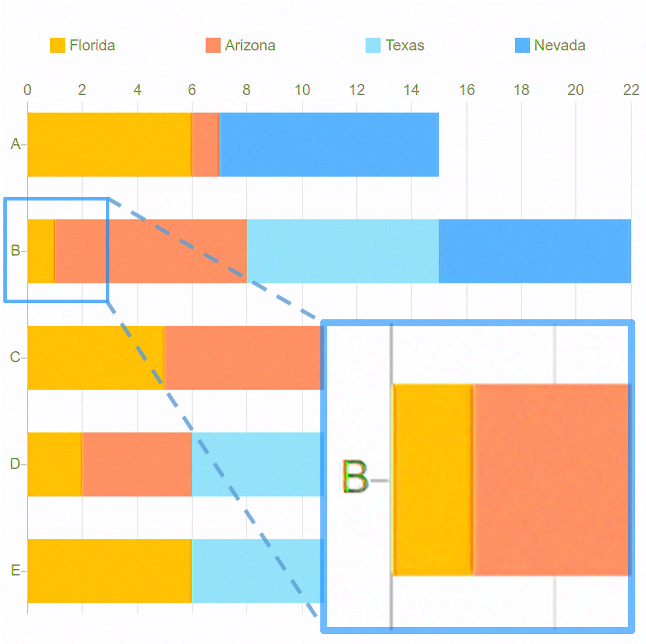}}
  \hfil
  \subfloat[StegaStamp]{\includegraphics[width=.225\linewidth]{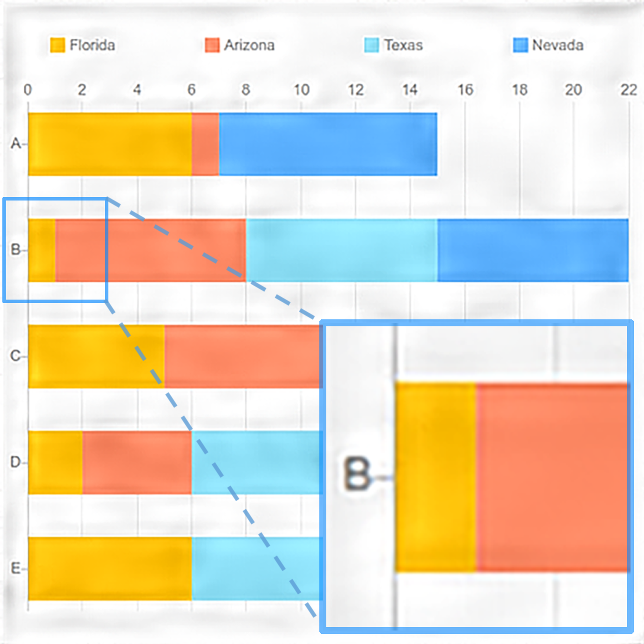}}
  \caption{\label{fig:qr_illu}Comparison of the encoded image quality with other methods.}
\end{figure}


\noindent \textbf{QR codes}
It is necessary to use an error-correction coding scheme when recovering secret information from the coded images.
Shannon’s~\cite{shannon1948mathematical} foundational work on coding theory for noisy channels has inspired many later studies. Reed et al.~\cite{reed1960polynomial} proposed a coding scheme based on a polynomial ring. Boneh et al.~\cite{boneh2002finding} utilized the Chinese remainder theorem to decode Reed–Solomon codes in polynomial time. BCH codes~\cite{bose1960class} are error-correction codes for binary group data. However, encoding binary data directly into carrier visual designs may result in obvious perturbations of the original appearance. ~\autoref{fig:qr_illu} presents an illustrative example comparing SteganoGAN~\cite{zhang2019steganogan} and StegaStamp~\cite{tancik2020stegastamp}. SteganoGAN~\cite{zhang2019steganogan} utilizes an adversarial training framework to embed arbitrary binary data into images based on Reed–Solomon codes. StegaStamp~\cite{tancik2020stegastamp} hides hyperlink bitstrings in photos using BCH codes. In ~\autoref{fig:qr_illu}, we can see that SteganoGAN may cause color variations in images with rich color content, while StegaStamp may generate visible noise when embedding information. Moreover, visualization charts usually have clean backgrounds and clear visual elements, which leads to a high decoding error rate in SteganoGAN. Due to these limitations, the two methods StegaStamp and SteganoGAN are not suitable for embedding information in data visualizations.

Barcodes are one of the most popular ways to transfer text information between computing devices. In our implementation, we convert arbitrary binary data provided by users into two-dimensional barcodes (QR codes) as our error correction coding scheme. QR codes~\cite{qrcodeweb} have essentially the same form as 2D images and therefore offer better visual effects than Reed–Solomon codes or BCH codes. QR codes have many advantages, such as a high capacity, a small printout size, and good dirt and damage resistance.
We refer the reader to the ISO/IEC 18004 specification~\cite{qrspec} for more technical details of QR codes. For our QR code dataset, we chose ``Binary'' as the information type and randomly generated 1500 QR code images with various character lengths from 1 to 2900 and various error correction capability levels of `Low' (`L'), `Medium' (`M'), `Quality' (`Q'), and `High' (`H'). Raising this level improves error correction capability, but also increases the size of QR code when embedding the same amount of data.


\begin{table}[bp]
\newcolumntype{M}[1]{>{\centering\arraybackslash}m{#1}}
\caption{Module Configurations of QR codes}
\centering
\small

\begin{tabular}{@{}M{2.6cm}M{2.6cm}M{2.6cm}@{}}

\toprule
Number of Characters & ECC Level & Resolution  \\
\midrule

 $ {\rm{1}} \sim {\rm{350}} $         & 'High'       & ${\rm{100}} \times {\rm{100}}$    \\
\midrule
 $ {\rm{351}} \sim {\rm{1000}} $         & 'Quality'       & ${\rm{200}} \times {\rm{200}}$    \\
\midrule
 $ {\rm{1001}} \sim {\rm{2000}} $         & 'Medium'       & ${\rm{200}} \times {\rm{200}}$    \\
\midrule

\multirow{1}{*} { $ {\rm{2001}} \sim {\rm{2900}} $ }
                      & 'Low'      & ${\rm{300}} \times {\rm{300}}$     \\

\bottomrule
\end{tabular}
\label{tab:table_qr}
\end{table}

\subsection{Visual Importance Map}
At a high level, the goal of the encoder network is to embed a large amount of information into a graphical chart while leaving the coded image perceptually identical to the original. A straightforward solution is to train a model to minimize the mean squared error (MSE) of the pixel difference between the original chart and the encoded chart~\cite{baluja2017hiding}. However, this metric is unreliable for assessing the visual quality of a chart~\cite{wang2004image}. For example, the encoder may generate visible noise in important areas since the MSE metric weights all pixels equally. To solve this problem, we introduce a visual importance map as a perceptual constraint to preserve the visual quality of the carrier chart image when embedding information.










\begin{table}[tbp]

\caption{Visual Importance Map Comparison}
\newcolumntype{M}[1]{>{\centering\arraybackslash}m{#1}}
\renewcommand\arraystretch{1.1}
\centering
\small

\begin{tabular}{@{}M{1.5cm}M{1.5cm}|M{1.1cm}|M{1.0cm}M{1.0cm}M{1.0cm}|@{}}

\bottomrule
\multicolumn{2}{|c|}{Method} & Dataset & $CC \uparrow$ & $RMSE \downarrow $ & ${R^2} \uparrow $ \\
\hline

\multicolumn{2}{|c|}{Bylinskii et al.~\cite{bylinskii2017learning}} & {\multirow{4}{*}{MASSVIS}} & {0.686} & {0.165} & {0.369} \\
\cline{1-2} \cline{4-6}

\multicolumn{1}{|c|}{\multirow{3}{*}{\textbf{Ours}}} & ${{\cal L}_{bce}}$ & {} & {0.866} & {0.123} & {0.715} \\
\cline{4-6}

\multicolumn{1}{|c|}{} & ${{\cal L}_{ssim}} $ & {} & {0.866} & {0.126} & {0.714} \\
\cline{4-6}

\multicolumn{1}{|c|}{} & ${{\cal L}_{bce}} + {{\cal L}_{ssim}} $ & {} & \textbf{0.868} & \textbf{0.117} & \textbf{0.720} \\
\toprule

\end{tabular}

\label{tab:table_vi}
\end{table}


Although many methods of saliency prediction have been proposed for natural images, the features of data visualizations are quite distinct. The visual importance of visualizations depends not only on the image context but also on higher-level factors (e.g., semantic categories)~\cite{bylinskii2017learning}. A visual importance map assigns different scores to different regions of an image, which can help to preserve the visual quality of the important content in the encoding stage.

Inspired by Bylinskii et al.~\cite{bylinskii2017learning}, we build a prediction model based on the MASSVIS dataset with eye movement annotations~\cite{borkin2015beyond}. Instead of leveraging an original fully convolutional network (FCN) architecture, we fine-tune a state-of-the-art salient object detection network, BASNet~\cite{qin2019basnet}.

As shown in~\autoref{fig:net_struc}(a), we use a network architecture the same as BASNet~\cite{qin2019basnet}, based on U-Net~\cite{ronneberger2015u} with a multiscale residual refinement module (RRM). Our visual importance network receives a three-channel RGB chart image with a resolution of $H \times W$ as input and outputs a one-channel $H \times W$ visual importance map. Similar to ResNet~\cite{he2016deep}, we use basic ResBlocks to obtain feature maps with different resolutions. Subsequently, we upsample the feature maps from the previous stage and concatenate them. The last feature map is then sent to the RRM for refinement. We refer the reader to the description of BASNet~\cite{qin2019basnet} for more details.

\begin{figure}[tbp]
  \centering
  \includegraphics[width=1.0\columnwidth]{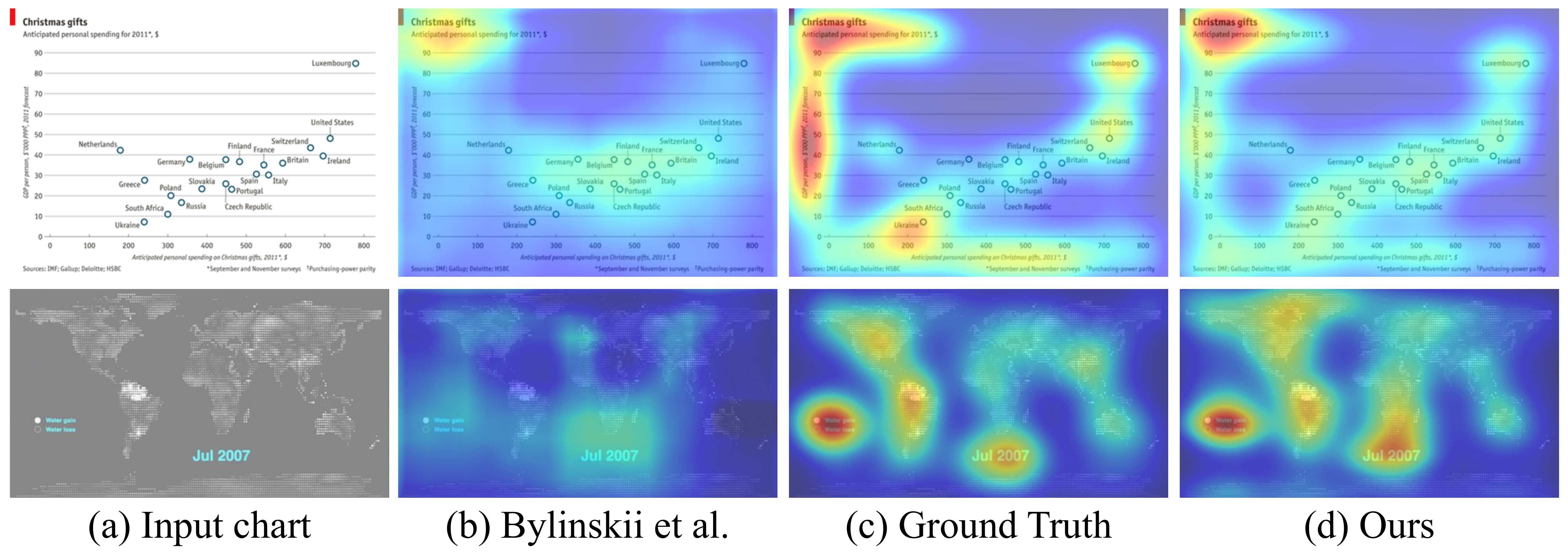}
  \vspace{-12pt}
  \caption{\label{fig:visim_illu}Visual importance maps predicted by our network are more similar to the distribution of ground truth.}
\end{figure}

The aim of the original BASNet is to detect salient objects, while our goal is to obtain a real-valued visual importance map describing the importance scores of all pixels. Our training process is based on the MASSVIS dataset~\cite{borkin2015beyond}.
Accordingly, to obtain more accurate high-level semantic context information and low-level details, we use a hybrid training loss:
\begin{eqnarray}
{{\cal L}_{prep}} = {{\cal L}_{bce}} + {{\cal L}_{ssim}}
\end{eqnarray}
where ${{\cal L}_{bce}}$ represents the BCEWithLogits loss~\cite{de2005tutorial} and ${{\cal L}_{ssim}}$ denotes the structural similarity index (SSIM) loss~\cite{wang2003multiscale}.

Given the ground-truth importance map at each pixel $p$, ${G_p} \in [0,1]$, over all pixels $p = 1,...,N$, the BCEWithLogits loss is defined as:
\begin{eqnarray}
{{\cal L}_{bce}}(G,V) =  - \frac{1}{N}\sum\limits_{p = 1}^N ( {G_p}\log {V_p} + {\rm{ }}(1 - {G_p})\log (1 - {V_p}))
\end{eqnarray}
where ${V_p}$ is the prediction of the visual importance network. The BCEWithLogits loss is defined in a pixelwise manner and facilitates the semantic segmentation of all pixels.

Let ${\rm{x}} = \{ {x_p}|p = 1,2,...,N\} $ and ${\rm{y}} = \{ {y_p}|p = 1,2,...,N\} $ be two corresponding image patches extracted from the ground-truth importance map ${G_p}$ and the predicted result ${V_p}$, respectively.
Let ${\mu _x}$ and ${\mu _y}$ be the means of ${\rm{x}}$ and ${\rm{y}}$, let $\sigma _x^2$ and $\sigma _y^2$ be their variances, and let ${\sigma _{xy}}$ be their covariance. The SSIM is defined as:
\begin{eqnarray}
\label{eqa:SSIM}
SSIM({\rm{x}},{\rm{y}}) = \frac{{(2{\mu _x}{\mu _y} + {C_1})(2{\sigma _{xy}} + {C_2})}}{{(\mu _x^2 + \mu _y^2 + {C_1})(\sigma _x^2 + \sigma _y^2 + {C_2})}}
\end{eqnarray}
where ${C_1} = {0.01^2}$ and ${C_2} = {0.03^2}$ are scalar constants. The SSIM loss is computed as ${{\cal L}_{ssim}} = 1 - SSIM({\rm{x}},{\rm{y}})$. This loss can help capture the structural information of the input chart image as the weights of the boundaries of visual elements increase.

The hybrid loss presented here helps us preserve the multiscale features of the visual importance map. 
Since we aim to get a visual importance map that is similar to the distribution of ground truth with smooth gradient for pixels, we discard IoU loss used in BASNet~\cite{qin2019basnet}.
For evaluation, we use the same metrics and test dataset as in the report by Bylinskii et al.~\cite{bylinskii2017learning}. Cross Correlation ($CC$) is a common metric used for saliency evaluation. Root-Mean-Square Error ($RMSE$) and the ${R^2}$ coefficient measure how correlated two maps are. 
We conduct an ablation study over different loss terms to demonstrate the effectiveness of the hybrid loss. The experimental results are shown in~\autoref{tab:table_vi}. Higher $CC$ values, lower $RMSE$ values and higher ${R^2}$ values are better. As we can see, the hybrid loss outperforms others. 
~\autoref{fig:visim_illu} demonstrates that the visual importance maps predicted by our network are more similar to the ground truth distribution than those of Bylinskii et al.~\cite{bylinskii2017learning}.

\subsection{Encoder and Decoder}
The encoder network is designed to embed information within a carrier chart image while minimizing perturbations in the encoded image. The decoder network recovers the embedded secret information from an encoded image.
Inspired by DeepSteg~\cite{baluja2017hiding}, we use a network structure similar to that of an autoencoder.
As shown in~\autoref{fig:net_struc}(b), the feature vectors of the input carrier chart image and the QR code image to be embedded are first extracted using $3 \times 3$ filters and $4 \times 4$ filters and the ReLU activation function~\cite{nair2010rectified}. Then, these two vectors are concatenated and fed to the next 3 convolutional layers, followed by batch normalization (BN)~\cite{ioffe2015batch} and a ReLU activation function. 
After upsampling with $4 \times 4$  filters (stride = 2) and a ReLU function, the resolution of the output encoded image is the same as that of the input chart image. Then we use two convolution layers with $3 \times 3$ filters and a $\tanh $ function to obtain the final output encoded image.

The architecture of the decoder network (~\autoref{fig:net_struc}c) is simpler than that of the encoder network. The encoded chart image is passed through a series of convolutional layers with $3 \times 3$ filters to produce the decoded QR code image as the final output. After error correction, we can obtain the reconstructed information in the same form as the original input message.

Different from DeepSteg~\cite{baluja2017hiding}, we introduce the visual importance map to supervise the encoder network. The encoder loss
is defined as:
\begin{eqnarray}
{{\cal L}_{Enc}} = V \odot {{\cal L}_{mse}} = \frac{1}{N}\sum\limits_{p = 1}^N ( {V_p} \odot {({I_{{c_p}}} - {I_{{c_p}}}^{'})^2})
\end{eqnarray}
Here, ${V_p}$ is the predicted visual importance map, which can be regarded as a weight matrix; regions with higher visual importance scores will have higher weights. ${I_c}$ is the input carrier chart image, and ${I_c}^{'}$ is the output encoded image. $N$ is the number of pixels.

The decoder network is supervised with a cross-entropy loss. Thus, we define the joint loss as:
\begin{eqnarray}
{{\cal L}_{joint}} = {{\cal L}_{Enc}} + \alpha {{\cal L}_{Dec}} = V \odot {{\cal L}_{mse}}({I_c},{I_c}^{'}) + \alpha {{\cal L}_{mse}}({I_s},{I_s}^{'})
\end{eqnarray}
where ${I_s}$ is the original input QR code image and ${I_s}^{'}$ is the reconstructed result of the decoder network. $\alpha $ is the weight of the decoder loss, which indicates a tradeoff between the visual quality of the encoded images and the decoding accuracy. Higher $\alpha $ results in higher decoding accuracy and decreased perceptual quality. $\alpha $ is set to $0.25$ in our implementation.

During the training process, we train the encoder network and decoder network simultaneously. That is, the joint loss incorporates both the visual quality of the coded image and the reconstruction error. 
In this study, our \emph{encoder-decoder} network needed a smaller structure and fewer parameters to fit the dataset since our visualization data available for training were limited compared with natural images. 
We trained our encoder and decoder networks using the aforementioned dataset consisting of data visualizations and QR codes. During training, the images were resized to $256 \times 256$ with antialiasing. We used the Adam optimizer~\cite{kingma2014adam} with an initial learning rate of $lr = 1{e^{ - 3}}$. We trained the encoder and decoder networks until the joint loss converged, without utilizing the validation set.

\begin{figure}[htb]
  \centering
  \subfloat[Original chart]{\includegraphics[width=.32\linewidth]{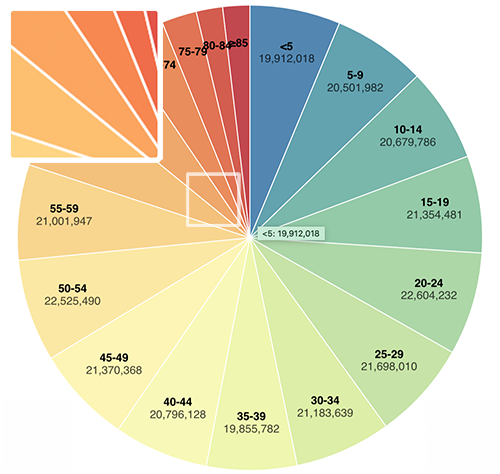}}
  \hfil
  \subfloat[Encoded chart]{\includegraphics[width=.32\linewidth]{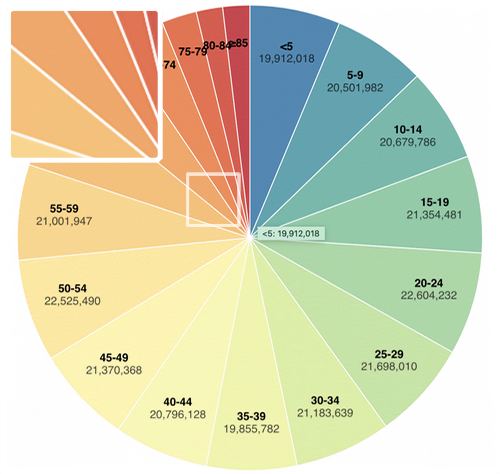}}
  \hfil
  \subfloat[Residual $\times 10$]{\includegraphics[width=.32\linewidth]{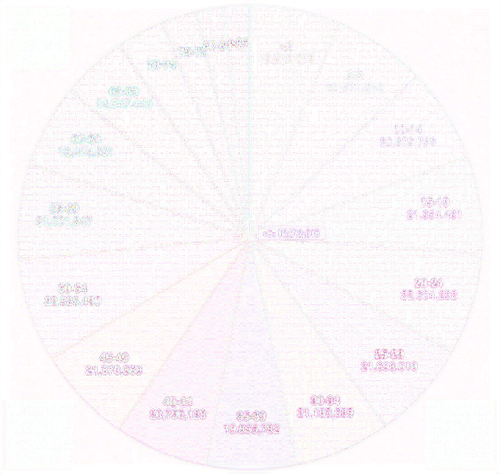}}
  \vspace{-5pt}
  \caption{\label{fig:enc_samp}Comparison of image quality before and after encoding.}
\end{figure}

In the testing process, the encoder and decoder can be used separately. \autoref{fig:enc_samp} shows an example of results obtained using the encoder. The coded image looks identical to the original chart image since the perturbations in the coded image are unobservable to human perception. We enhance the intensity of the residuals (\autoref{fig:enc_samp}c) between the input chart image (\autoref{fig:enc_samp}a) and the output coded image (\autoref{fig:enc_samp}b) by a factor of 10 times to more clearly show the differences. The results demonstrate that our VisCode model can embed information while preserving the perceptual quality of the original chart image.

\begin{figure*}[tbp]
\centering
\includegraphics[width=1.0\linewidth]{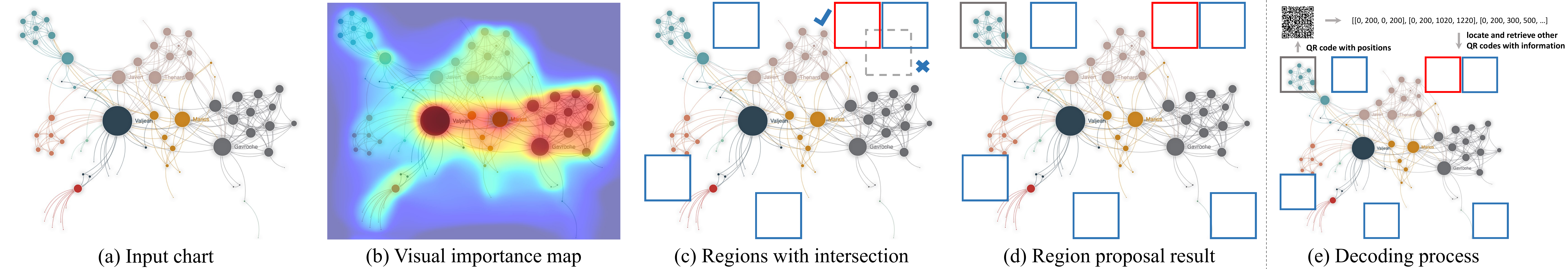}
\vspace{-12pt}
\caption{\label{fig:pos_illu}Ilustration of the embedding region proposal algorithm. We localize the optimal embedding regions based on the visual importance map and avoid overlaps of the proposal region boxes. The position QR code makes the decoding process more efficient and accurate. }
\end{figure*}

\subsection{Large-Scale Data Embedding}
\label{subsec:LSDE}
Our VisCode system is designed to take a chart image with a resolution of $H \times W$ and arbitrary binary data as input, making it suitable for various actual scenarios.
Embedding a large amount of information in an image is a challenging problem because it is easy to generate obvious noise or cause deviations in color or texture. To overcome this challenge, Wu et al.~\cite{wu2018stegnet} proposed learning the conditional probability distribution of the original image. However, this method may affect viewers' understanding of a chart since semantic information such as color maps and data points are particularly meaningful in data visualizations. Accordingly, we propose a novel approach for the large-scale embedding of information in regions to which the human visual system is insensitive.

Although arbitrary binary data can be accepted as input information, we focus on a text message to facilitate the explanation. Suppose that the user inputs text with a length of $Le{n_T}$ characters. We first split the text into several blocks:
\begin{eqnarray}
\label{eqa:numb}
Nu{m_B} = \left\lceil {Le{n_T}/\eta } \right\rceil ,\;Le{n_B} = \left\{ \begin{array}{l}
Le{n_T}/Nu{m_B},\;other\;blocks\\
Le{n_T}\bmod Nu{m_B},\;last\;block
\end{array} \right.
\end{eqnarray}
where $Nu{m_B}$ denotes the number of blocks and $Le{n_B}$ denotes the length of each block. Considering that there are various module configurations available for QR codes, which may influence the effects of \emph{the encoder-decoder} network, we conducted tests with various mapping relations between the length of the text and the module configurations of QR codes. ~\autoref{tab:table_qr} shows the results. We use these mapping relations as criteria instead of performing dynamic calculations, which reduces the search time. The parameter $\eta $ can be specified in accordance with the user's preference. A higher $\eta $ corresponds to more text in each QR code. We set this parameter to $800$ by default.

\begin{algorithm}[htb]
  \caption{ Embedding region proposal algorithm.}
  \label{alg:LSDEa}
  \begin{algorithmic}[1]
    \Require
    $V$: the visual importance map; $Text$: the user input text; $\eta $: the acceptable maximal number of characters in a single QR code;
    \Ensure
    ${R_{{o\_{set}}}} = \{ {R_{{o_1}}},{R_{o_2}},...,{R_{{o_k}}}\}$: a set of optimal embedding regions;
    \State $ (W,H) = size(V) $, 
    \State $Calculate \ Nu{m_B}$ and $Le{n_B}$ based on ~\autoref{eqa:numb}
    \State ${R_{{o\_{set}}}} = \{ \}$, $cnt = 0$
    \State $kernel\_size = map(Le{n_B})$ based on ~\autoref{tab:table_qr}
    \State ${Full_{set}} \leftarrow Apply \ average \ pooling \ on \ V$
    \State sort$({Full_{set}})$ in increasing order
    \For{ $c\_region$ \textbf{in} ${Full_{set}}$}
    \For{ $s\_region$ \textbf{in} ${R_{{o\_{set}}}}$}
    \If{$Intersection(c\_region,s\_region) > 0$}
    \State $ conflict = true$
    \EndIf
    \EndFor
    \If{$ conflict = false \ and \ cnt <= Nu{m_B}$}
    \State ${R_{{o\_{set}}}}$.append($c\_region$), $cnt = cnt + 1$
    \EndIf
    \EndFor
    \label{code:filter} \\
    \Return ${R_{{o\_{set}}}}$
  \end{algorithmic}
\end{algorithm}

After determining the resolution of each QR code, we attempt to localize the optimal regions of the carrier chart image in which to embed these QR codes. A straightforward solution is to randomly select several regions, but this may create discontinuities near the boundaries between modified and unmodified regions that will look disharmonious.
Another brute-force solution is to list all possible regions and evaluate the effects of the \emph{encoder-decoder} network on each one. However, this may incur a substantial time cost due to the large search space.
Instead, inspired by region proposal techniques from the field of object detection (e.g., selective search~\cite{girshick2014rich} and RPN~\cite{ren2015faster}), we generate bounding boxes based on the visual importance map instead of the original chart image.

The corresponding problem can be formulated as follows. Given a chart image with a resolution of $H \times W$ and a text with a character length of $Le{n_T}$, we wish to find the $Nu{m_B}$ optimal regions with the smallest values in the visual importance map. The size of each region box depends on the mapping relations shown in ~\autoref{tab:table_qr}. We generate $Nu{m_B}$ QR code images containing the $Nu{m_B}$ blocks of text to be embedded.
Given the visual importance map ${{V_p}}$ and the size of each region box, we use average pooling to calculate a value for each region box:
\begin{eqnarray}
{R_v} = \frac{1}{{{N_b}}}\sum\limits_{p = 1}^{{N_b}} {{V_p}}
\end{eqnarray}
where ${{N_b}}$ is the kernel size of the filter, which we regard as the size of the region box. Then, we sort the region boxes in increasing order of their values to obtain the list of candidate regions. We preserve the top left region box $[(0,0),({x_{rd}},{y_{rd}})]$ for embedding the position information of the other $Nu{m_B}$ region boxes. A QR code containing the corresponding position information is also generated. To avoid overlap of the region proposal boxes, we adopt nonmaximum suppression (NMS)~\cite{neubeck2006efficient} based on their intersection values. \autoref{fig:pos_illu} shows the key steps of the embedding region proposal algorithm.
The detailed region proposal algorithm is presented in Algorithm~\ref{alg:LSDEa}. After determining the optimal regions, we crop the corresponding $Nu{m_B}+1$ patches of the input chart image and feed them into the encoder network with the $Nu{m_B}+1$ QR codes. In the decoding stage, we first extract the top left QR code with the position information from the coded chart image. Then, we locate other corresponding image patches and send them to the decoder network to retrieve the complete information. We utilize the position QR code instead of generating the visual importance map, which is more efficient and accurate.


\section{Applications}
\label{sec:apps}

VisCode can be used for a number of applications, such as embedding metadata or source code in visualization design and visualization retargeting.
Experiments related to the envisioned applications were implemented on a PC with an Intel Core i7 CPU, an NVIDIA GeForce 2080 Ti GPU, and 32 GB of memory. The visualization images were generated using D3~\cite{bostock2011d3} and Echarts~\cite{li2018echarts}. The deep learning framework was implemented based on Pytorch~\cite{paszke2019pytorch}.

\subsection{Metadata Embedding of Visualization Design}
Information visualization technology uses computer graphics and image processing techniques to convert data into graphics or images, allowing users to observe and understand data patterns more intuitively. Related to the processes of visualization design, display, and dissemination, there is usually considerable information that needs to be stored, such as the name of the designer, the design institution, the design time, the URL of the visualization web page, and historical revision information. Such related information for a visualization image is called metadata~\cite{greenberg2005understanding}.

\begin{figure}[htb]
  \centering
  \includegraphics[width=0.74\linewidth]{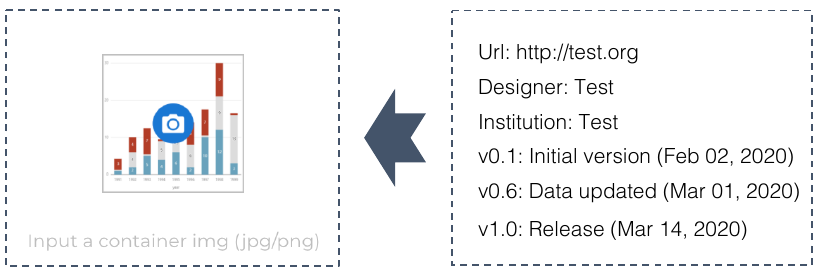}
\vspace{-10pt}
\caption{Application interface for embedding information in a visualization image.}
  \label{fig:case01}
\end{figure}

VisCode provides strong support for information steganography in visualization design. First, our information encoding method is implicit and does not affect the original visual design. \autoref{sec:eva} will demonstrate the efficiency of our encoding method. Second, the proposed approach has a fast encoding speed and a high decoding success rate. Therefore, it is possible to effectively encode and decode the author information of visual works, the source website, and other such information.

\autoref{fig:case01} shows the application interface for embedding information in a visualization image. The input to the VisCode system consists of a visualization image and the corresponding metadata. These metadata may include website links, author names, institutions, and revision logs. The metadata are hidden in the visualization image. Obtaining the metadata through the VisCode decoding process is simple. Using this application, the designer of a visual chart can hide common text information in the chart. On the one hand, the design work is implicitly protected, and on the other hand, there is no need to maintain additional modification log files corresponding to the image.
The ability to hide URL links in a chart also provides a convenient way to allow users to view web-based or interactive visualizations. As shown in~\autoref{fig:case01}, after the author hides the URL information in the image, a user who has obtained the image can use the decoding module of VisCode to obtain the URL and access the real-time interactive visualization corresponding to the static image.

\subsection{Source Code Embedding in a Visualization}
\begin{figure}[htb]
  \centering
\subfloat[Interaction]{\includegraphics[height=.25\linewidth]{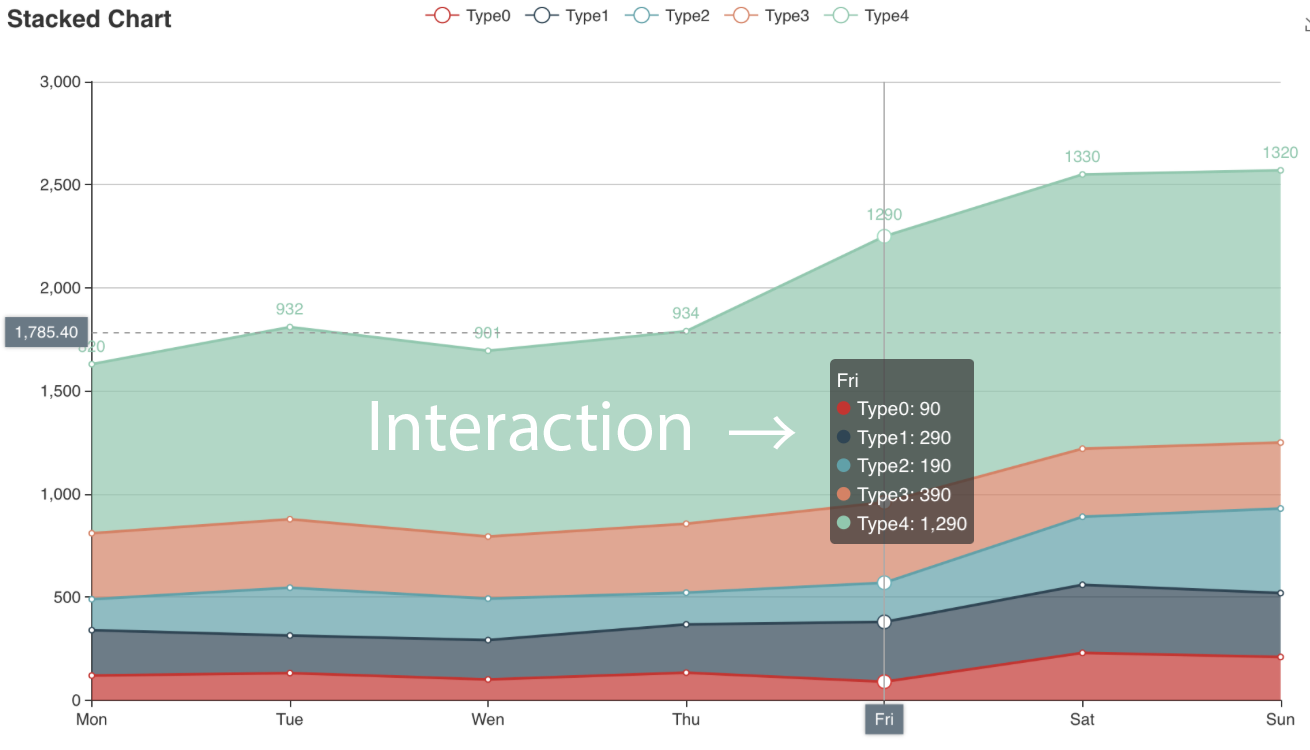}}
\hfil
\subfloat[Low resolution]{\includegraphics[height=.25\linewidth]{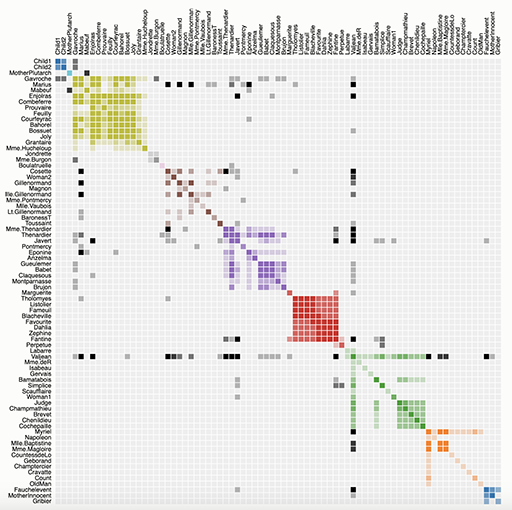}}
\hfil
\subfloat[3D]{\includegraphics[height=.25\linewidth]{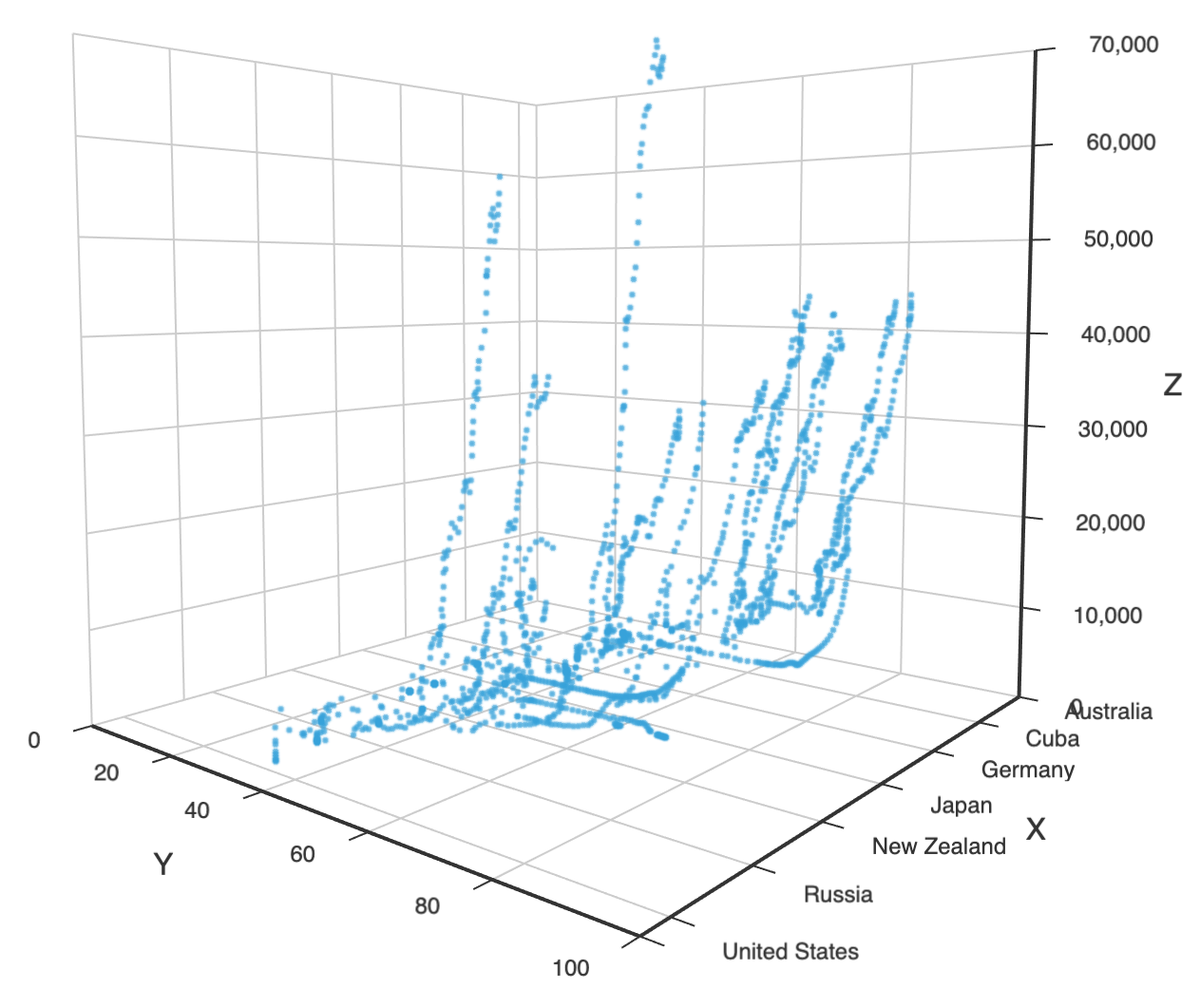}}
\hfil
\vspace{-5pt}
\subfloat[Animation (three frames)]{
\includegraphics[height=.20\linewidth]{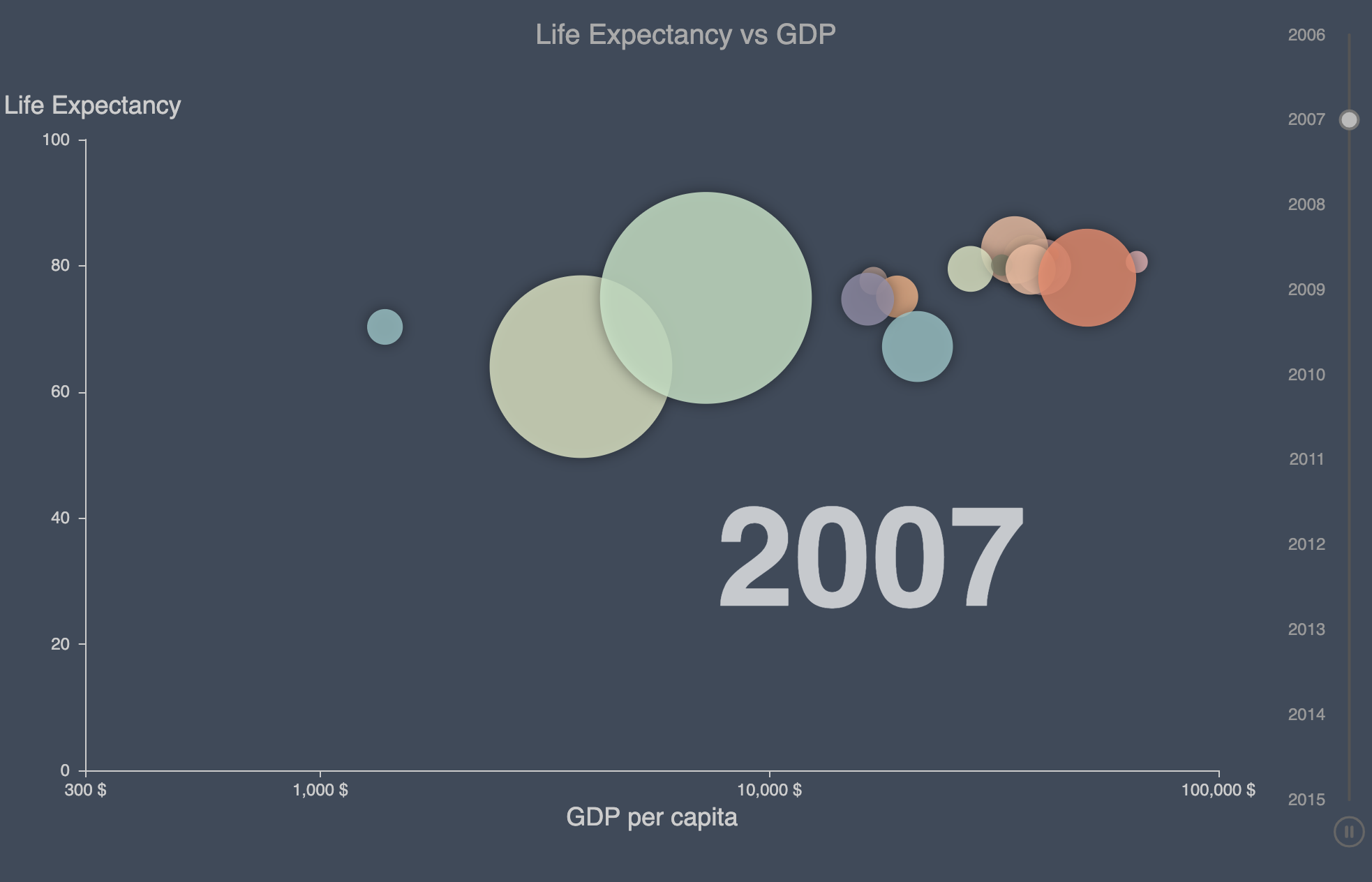}
\includegraphics[height=.20\linewidth]{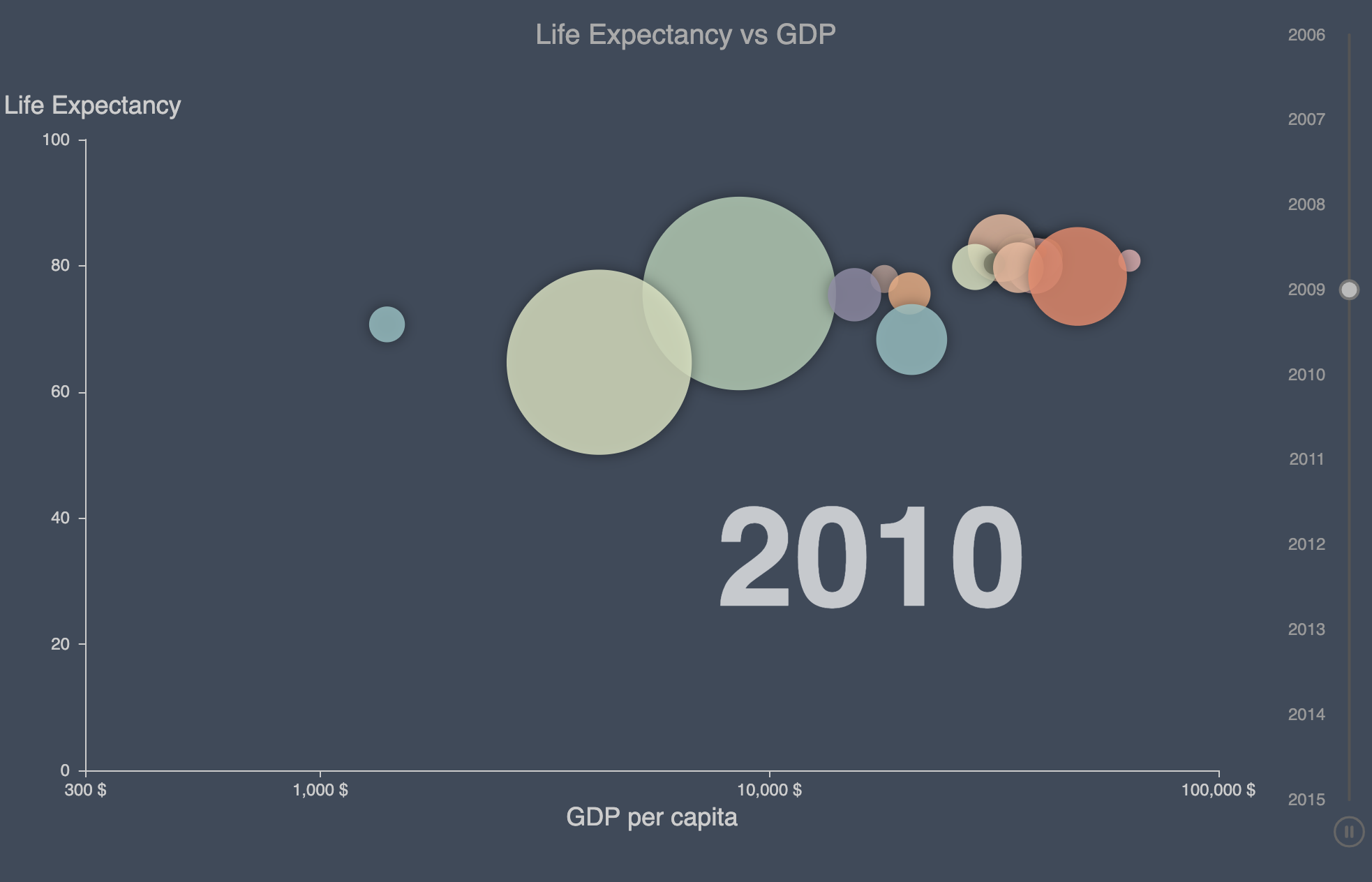}
\includegraphics[height=.20\linewidth]{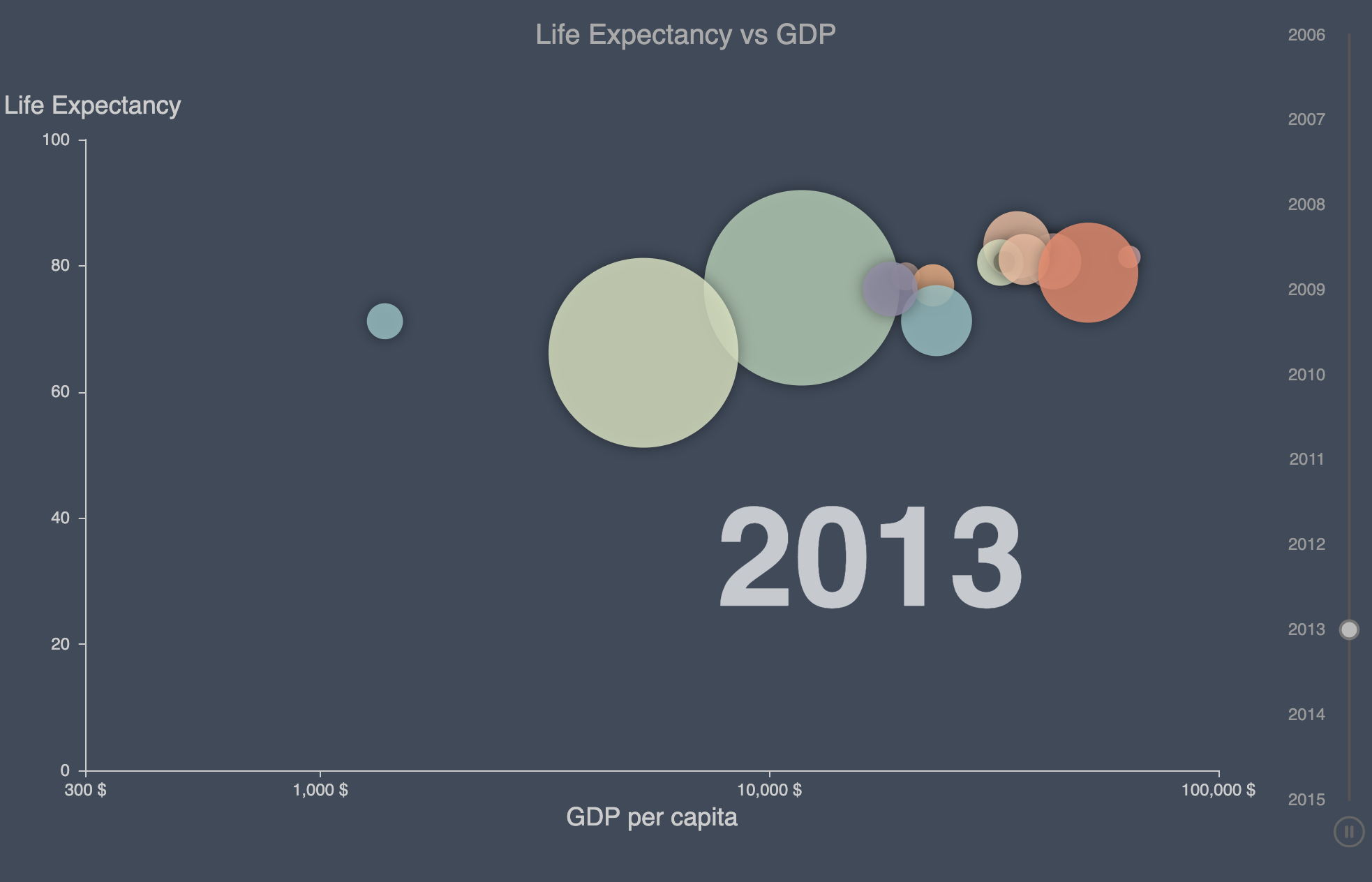}
}
\vspace{-5pt}
\caption{Three application scenarios involving the embedding of source code in visualizations.
}
\label{fig:case02_motivations}
\end{figure}

Due to the limitations of network connections and Web servers, the presentation of dynamic web pages is not as flexible as that of static images. Therefore, the display of static images is a very important means of sharing information visualizations. However, static image display has many shortcomings compared with web-based visualization applications. First, a mature information visualization toolbox usually provides multiple interaction modes to help users further explore the contents of data, such as time period selection, data area selection, map level selection in map visualizations, and clicking and dragging operations.
\autoref{fig:case02_motivations} presents several cases of source code embedding in visualizations. The interactive information label display shown in~\autoref{fig:case02_motivations}(a) is a frequently used function for information visualization tasks. Second, since static images are mostly stored in bitmap format, when the resolution of an image is not high, it cannot clearly present information, as shown in~\autoref{fig:case02_motivations}(b).
Third, using the 3D form is a convenient way for the user to observe the patterns in different views. This is not possible with a static visualization image as shown in~\autoref{fig:case02_motivations}(c).
Fourth, in a visualization involving temporal data, animation is commonly used to visualize data in different periods represented by time steps, as shown in~\autoref{fig:case02_motivations}(d).
Hans Rosling's dynamic visualization of population and income throughout the world is a representative example~\cite{maxmen2016three}.

To extend the usage of VisCode, we present an application scenario that provides users with a robust source code embedding function. In this source code embedding scenario, we encode the source code data into a static visualization image.
Because the VisCode framework supports very low loss of encoding quality and the encoding of large datasets, it can well support the encoding of source code in information visualizations. For the implementation of this application scenario, we first integrated common information visualization frameworks (such as D3~\cite{bostock2011d3} and Echarts~\cite{li2018echarts}) into the application and then built a functionality for encoding personalized visualization code into a specified image.
When a user inputs an encoded image, the system decodes the source code from the image and dynamically generates and displays the visualization file, usually in the form of a web page.

\begin{figure}[htb]
  \centering
  \includegraphics[width=1.0\linewidth]{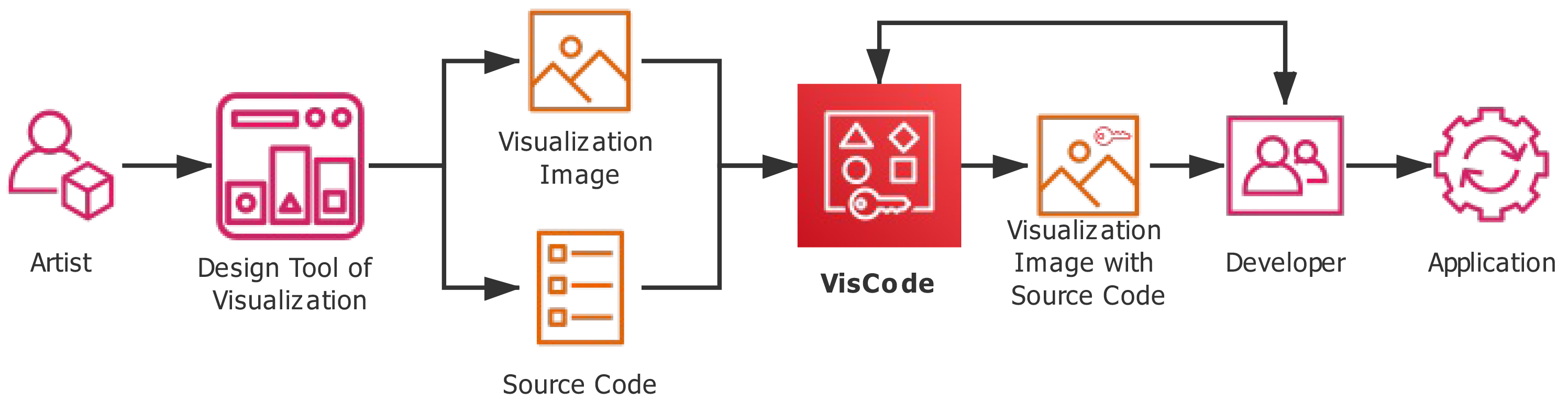}
\vspace{-14pt}
\caption{A pipeline for collaboration between an artist and a developer using the VisCode system.}
  \label{fig:case02_pipeline}
\end{figure}

In addition, source code embedding through VisCode can facilitate collaboration between artists and developers working on visualization projects. In the production processes of large enterprises, artists usually use existing visualization design tools to create visualizations based on aesthetic considerations and then provide the renderings and source code generated by those design tools to developers. Because the interactions between these two roles are frequent and the visual design may undergo many revisions, it can be easy to confuse the design images and codes from different versions. The need to maintain or reconstruct the correspondence relationship between code files and images will increase the difficulty of file sharing. As an alternative approach, VisCode can effectively convert an image and its corresponding source code for a design into a single static image, thus reducing the occurrence of low-level errors. \autoref{fig:case02_pipeline} shows a pipeline for collaboration between an artist and a developer using the VisCode system. The artist can encode the source code into a static visualization image, and the developer can decode the source code and use it in a visualization application.

\begin{figure}[htb]
  \centering
\subfloat[Input]{\includegraphics[width=.198\linewidth]{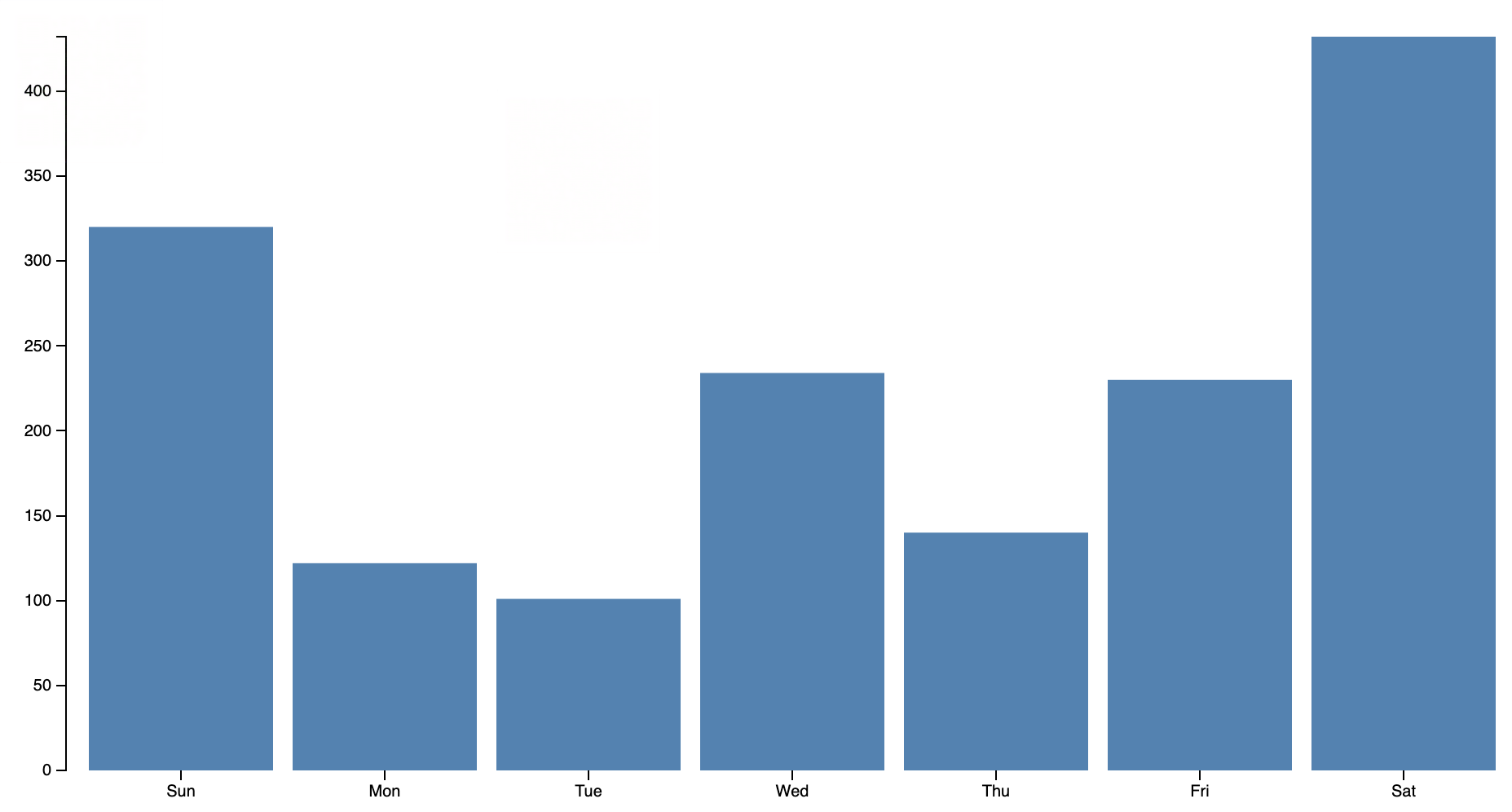}}
\hfil
\subfloat[Ouput 1]{\includegraphics[width=.198\linewidth]{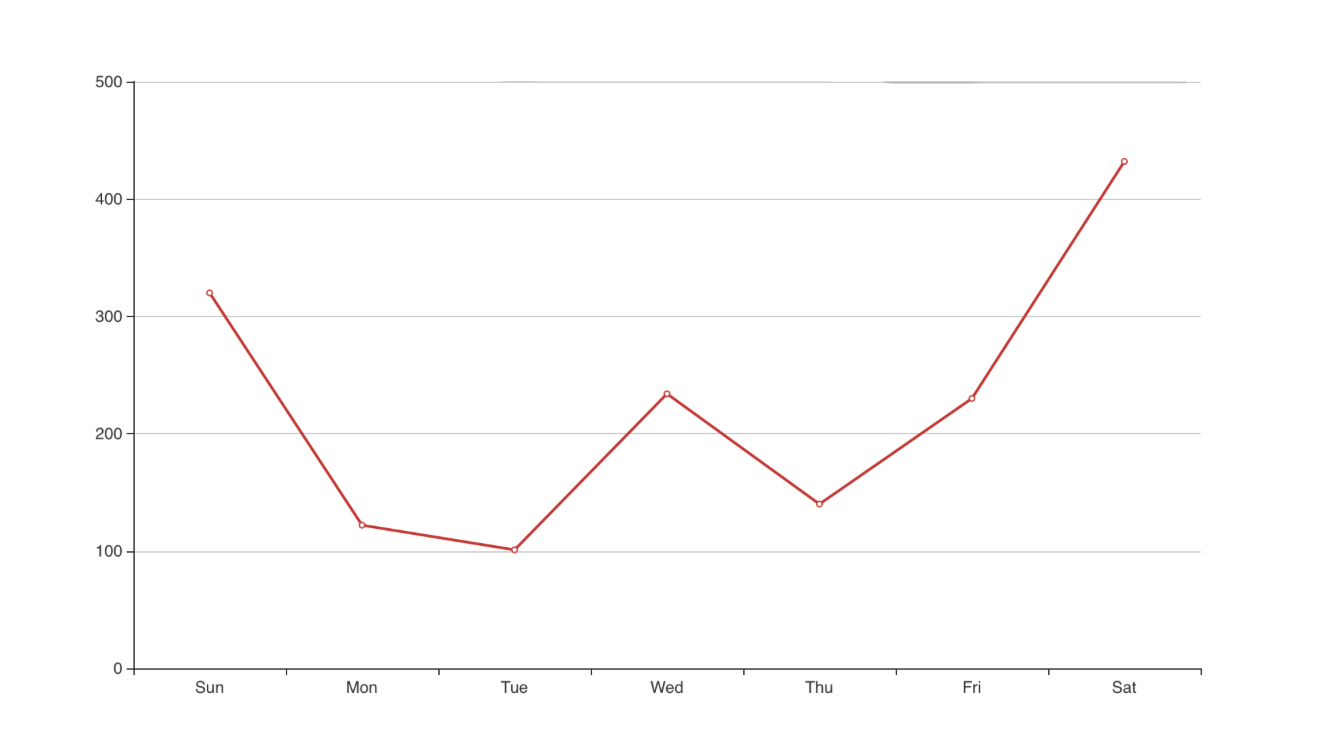}}
\hfil
\subfloat[Ouput 2]{\includegraphics[width=.198\linewidth]{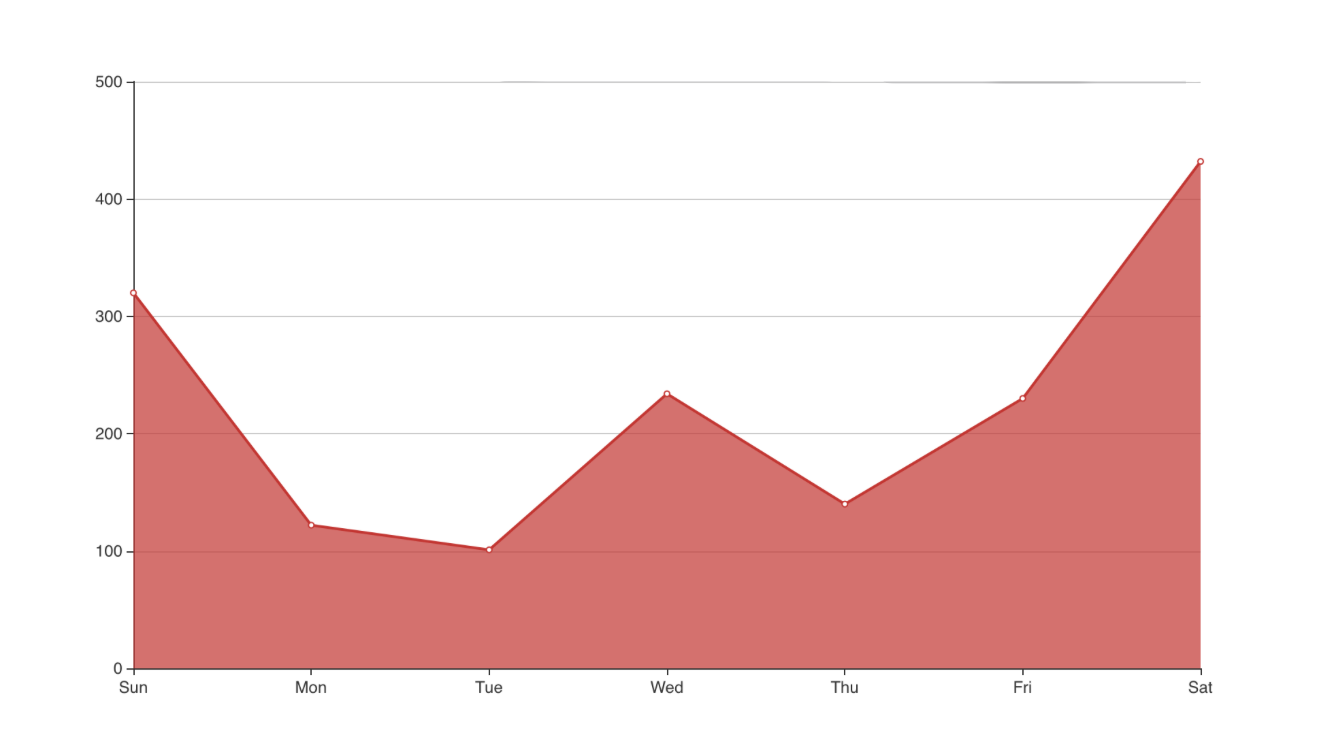}}
\hfil
\subfloat[Ouput 3]{\includegraphics[width=.198\linewidth]{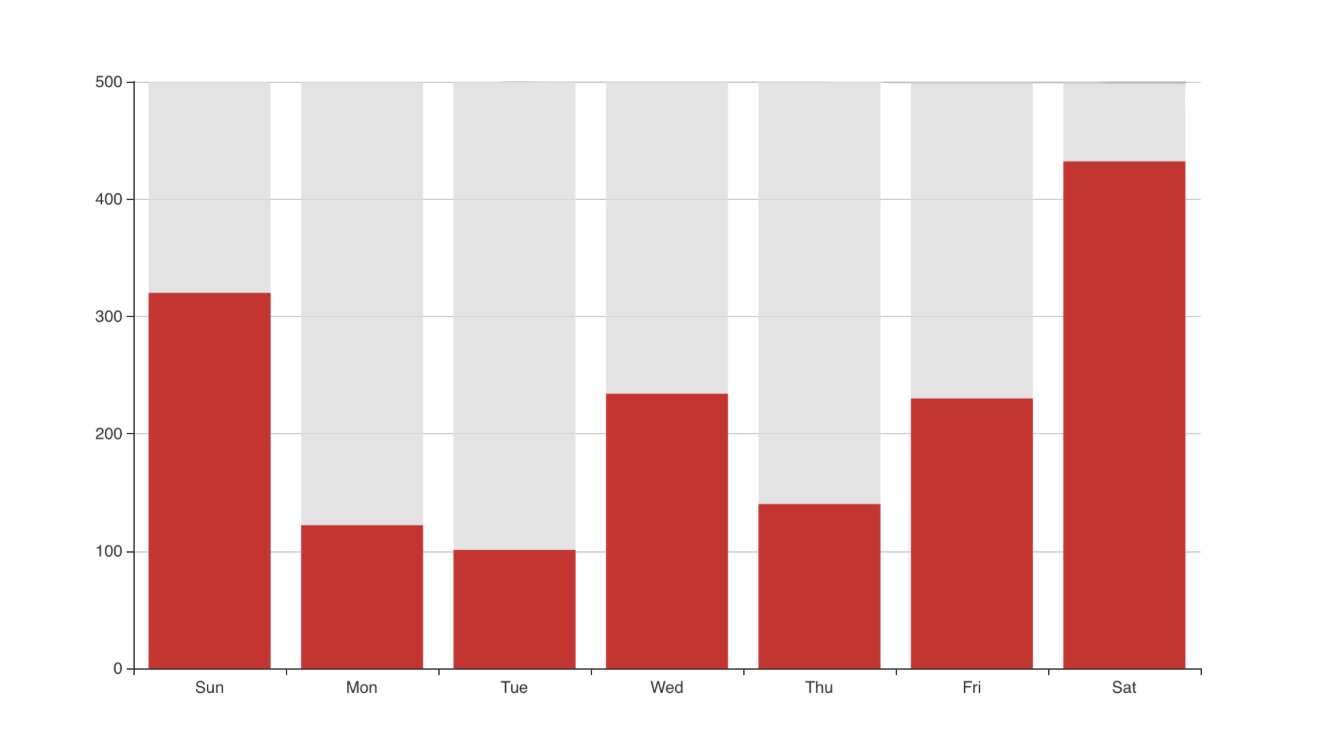}}
\hfil
\subfloat[Ouput 4]{\includegraphics[width=.16\linewidth]{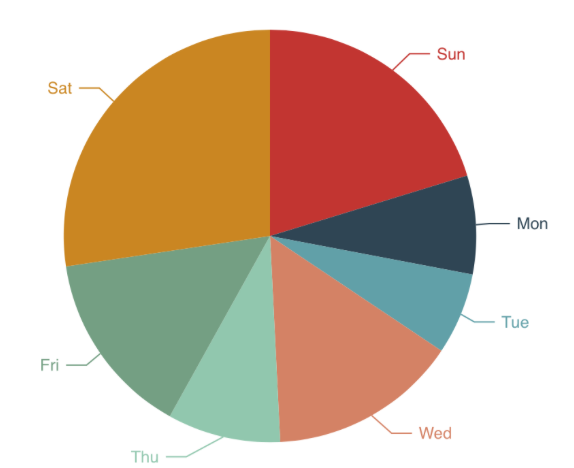}}
\vspace{-5pt}
\caption{Retargeting of the form of representation of information. The input visualization can be converted to other visualization forms.}
  \label{fig:case03_retar_rep_type}
\end{figure}

\subsection{Visualization Retargeting}
In addition to source code embedding, another application of VisCode is to retarget a visualization. With the development of big data and deep learning technologies, visualization retargeting based on pattern recognition has gained the attention of researchers in recent years, as in the work of Poco et al.~\cite{poco2017extracting}. Visualization retargeting is very useful for the creation of visualizations.
On the one hand, it reduces the designer's workload. On the other hand, it broadens the artist's creative space.
For some common chart types, existing research has yielded methods of extracting data from a visualization image and retargeting the visualization.
For example, Poco et al.~\cite{poco2017extracting} presented an approach for identifying values, legends, and coordinate axes from a bar chart. Similar work has also been applied for information extraction and retargeting based on color density maps. 
However, for many types of visualizations, such as network diagrams, such recognition is still difficult. Visualization retargeting via pattern recognition is not feasible for some
visualizations if the visualization process is not irreversible. For example, even if the regions of a density map can be accurately identified, the original scattered data constituting the density map cannot be recovered.

\begin{figure}[htb]
  \centering
\subfloat[Encoded image]{\includegraphics[width=.445\linewidth]{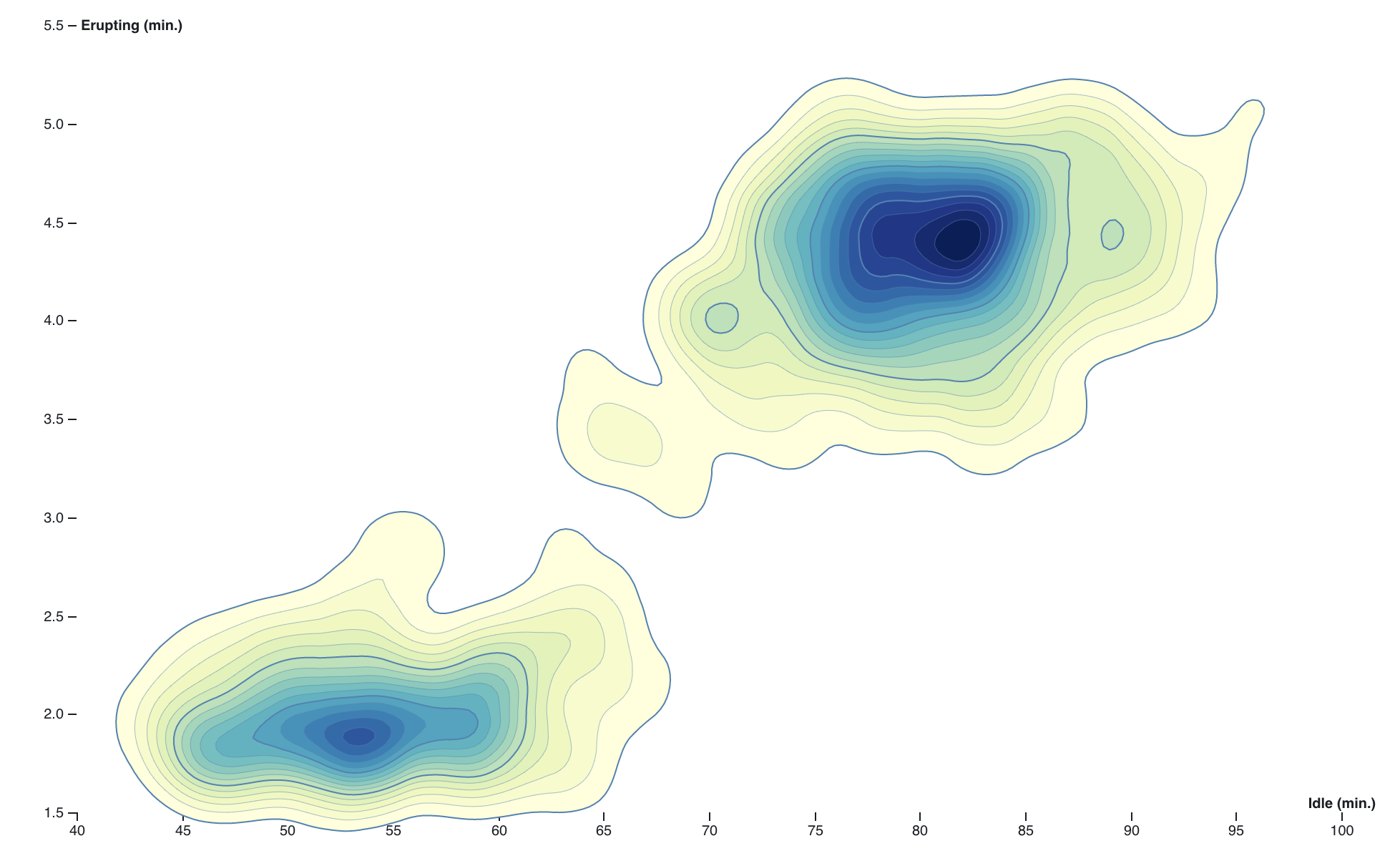}}
\hfil
\subfloat[Decoded scatter points]{\includegraphics[width=.445\linewidth]{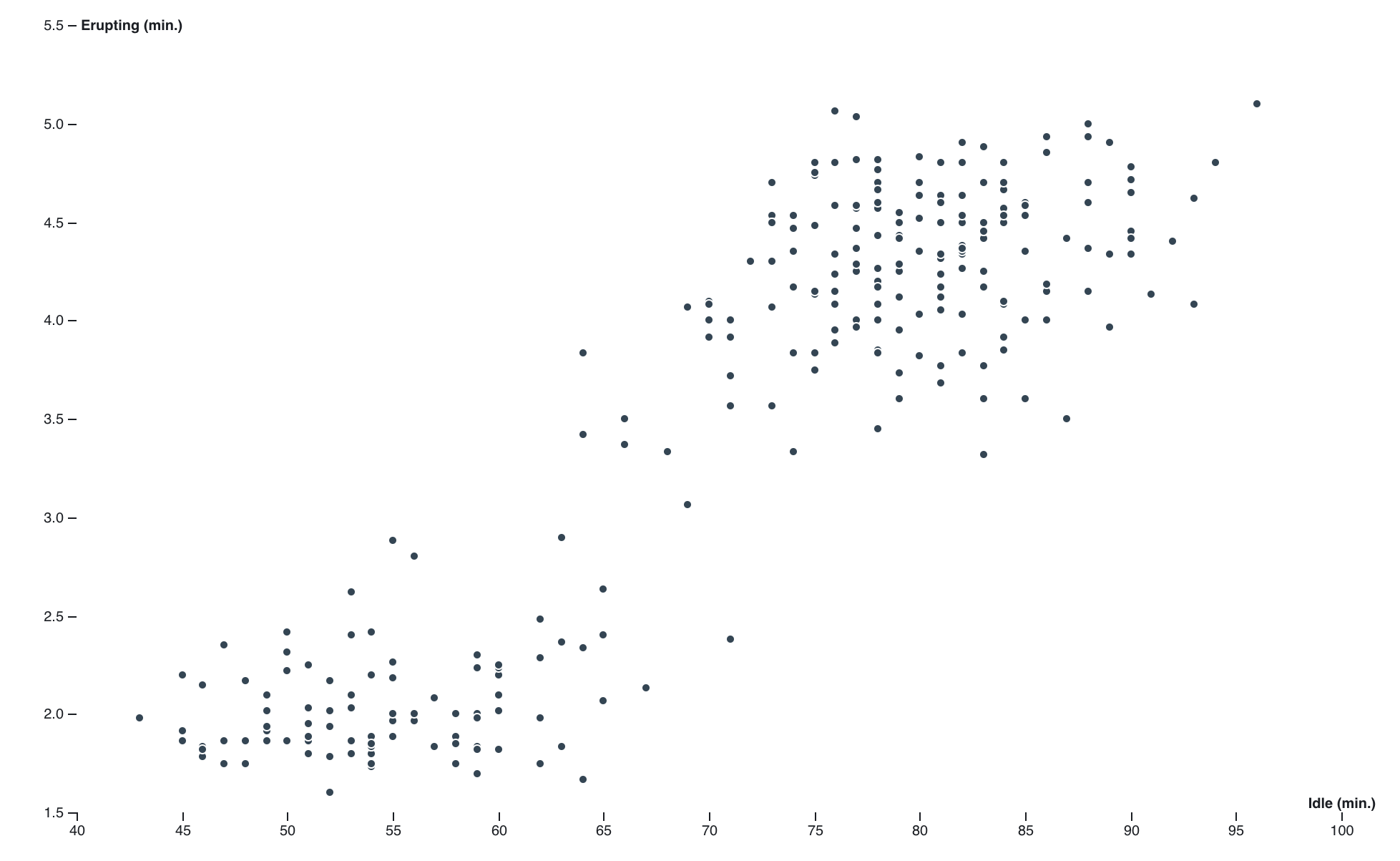}}\
\subfloat[Retargeting result 1]{\includegraphics[width=.445\linewidth]{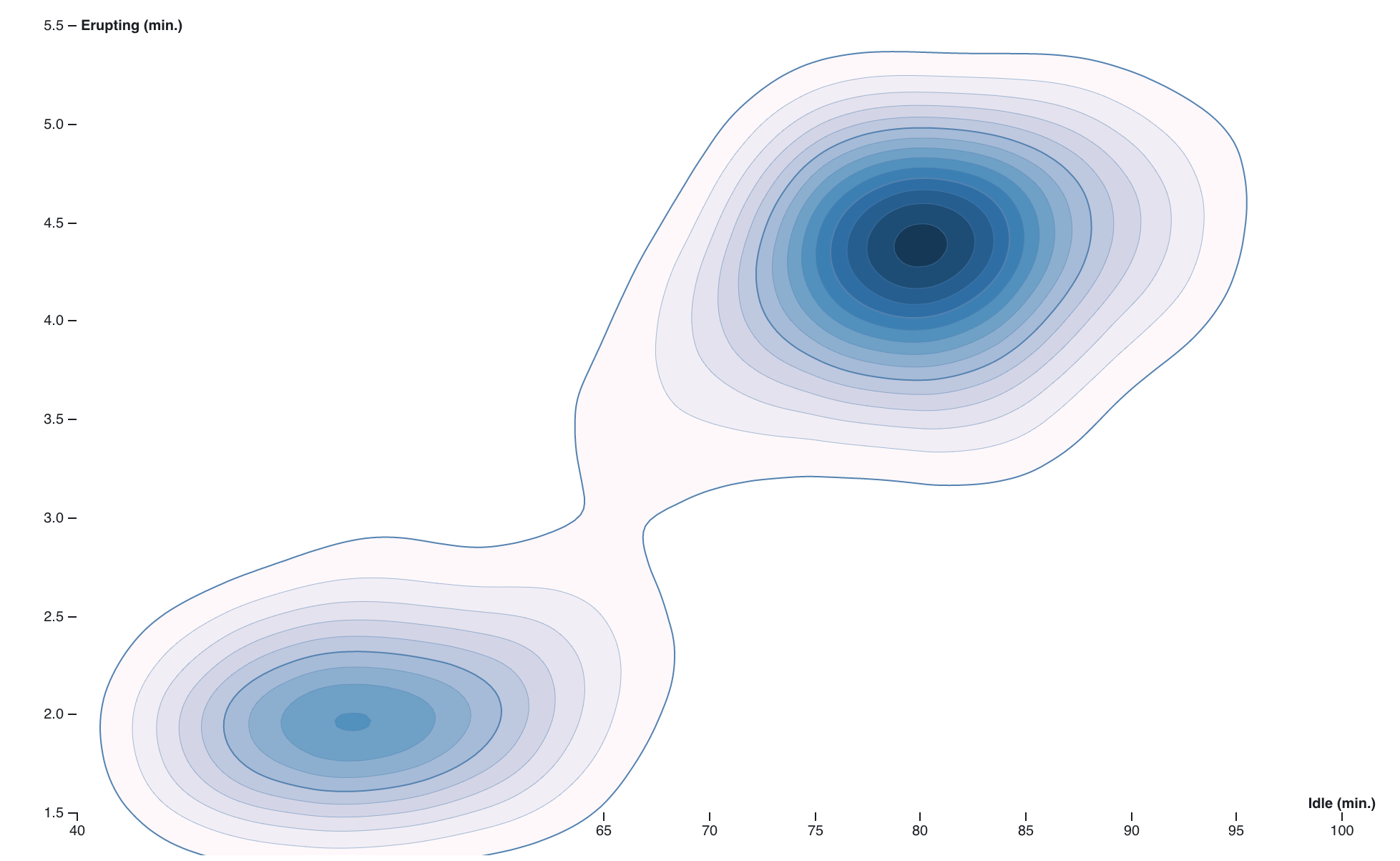}}
\hfil
\subfloat[Retargeting result 2]{\includegraphics[width=.445\linewidth]{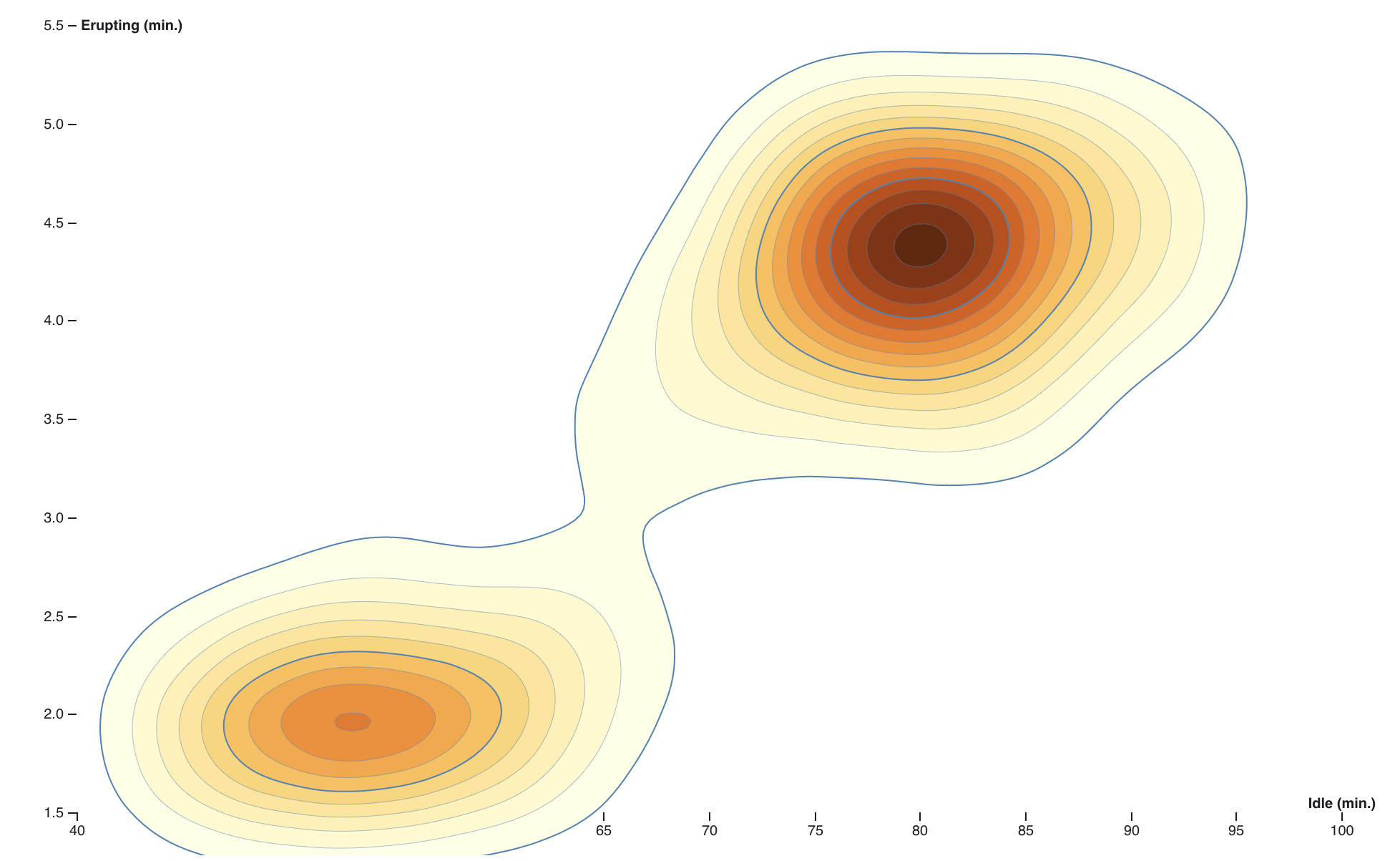}}
\vspace{-3pt}
\caption{Retargeting the visualization of a color density map. Scatter points are extracted from the input image and new colorful density maps are generated through the kernel density estimation.}
  \label{fig:case03_heatmap}
\end{figure}

However, if VisCode is used to encode the original data into a static image, then for a visualization that is saved in a certain format, such as JSON, high-quality visualization retargeting can be achieved without pattern recognition. Our implementation includes representation retargeting and theme retargeting. First, we encode the original data of the visualization and the visualization type into a static visualization image in JSON format. After decoding, the data are displayed in a visualization of the specified type. For users, the input is an encoded static image, and the output is a visual effect of different styles, as shown in~\autoref{fig:case03_retar_rep_type}. We define this form of retargeting as representation retargeting. Second, we can implement different color themes for users to choose based on an existing representation. This capability of color theme switching is called theme retargeting.
This application of VisCode can provide rich visual styles and color themes. It can also be used to enable deformations for the visualization of complex network diagrams from which information is difficult to extract, as shown in~\autoref{fig:pipeline}(a). From an encoded color density map, the original scattered data can be extracted through the decoding function of VisCode. Based on these scattered data and the kernel density algorithm~\cite{silverman1986density}, a new density map with a personalized bandwidth can be generated. \autoref{fig:case03_heatmap} presents an example of the retargeting of a visualization of a color density map. \autoref{fig:case03_heatmap}(a) shows the input static image. \autoref{fig:case03_heatmap}(b) shows the extracted scattered data, and \autoref{fig:case03_heatmap}(c-d) show two personalized color density maps.

\section{Evaluation}
\label{sec:eva}
We evaluate our VisCode model from three aspects: \emph{steganography indices}, the evaluation of encoded image quality with the different number of message bits; \emph{steganography defense}, the evaluation of decoding accuracy in various corruption scenarios through digital transmission; and \emph{time performance}, the evaluation of embedding information and recovery time of different settings.

\subsection{Steganography Indices}
We used three metrics to evaluate the performance of our VisCode model: the peak signal-to-noise ratio (PSNR)~\cite{almohammad2010stego}, the SSIM~\cite{wang2004image}, and the learned perceptual image patch similarity (LPIPS)~\cite{zhang2018unreasonable}. The PSNR~\cite{almohammad2010stego} is a widely used metric for evaluating image distortion. We computed the PSNR by comparing the coded image ${I_c}^{'}$ and the original chart image ${I_c}$, each consisting of $N$ pixels:
\begin{equation}
\begin{aligned}
PSNR = 20 \cdot {({\log _{10}}{P_{\max }} - {\log _{10}}\frac{1}{N}\sum\limits_{p = 1}^N ( {I_{{c_p}}} - {I_{{c_p}}}^{'})^2})
\end{aligned}
\end{equation}
where ${{P_{\max }}}$ is the maximum possible difference between two pixel values, which we set to $1.0$ in our experiment since the RGB values of our images are ranged from $0$ to $1$.

Because the PSNR measures the distance between each pair of pixels independently, we also considered the SSIM to measure the structural similarity between two images. The SSIM is defined as shown in Equation~\ref{eqa:SSIM}. It is a patch-level perceptual metric. However, whereas both the PSNR and SSIM are low-level metrics, Zhang et al.~\cite{zhang2018unreasonable} have demonstrated that perceptual similarity is influenced by visual representations instead of being a special function. Accordingly, the LPIPS, which relies on deep features, can better reflect human judgments.

We construct a test dataset of 500 chart images with various resolutions and types, as mentioned in ~\autoref{subsec:DV_dataset}.  
We compare our method with StegaStamp~\cite{tancik2020stegastamp} and SteganoGAN~\cite{zhang2019steganogan}. However, there are some limitations to these two methods. StegaStamp~\cite{tancik2020stegastamp} can embed only $100$ bits data. SteganoGAN~\cite{zhang2019steganogan} can embed small-scale data, but it will cost huge time when the number of bits increases to a threshold. Therefore, the two methods StegaStamp and SteganoGAN are not suitable for embedding large-scale information in data visualizations. 
For a fair comparison, we resize the chart images to the same resolution, and embed the same numbers of data bits into the test dataset. Higher PSNR, higher SSIM, and lower LPIPS are better. As shown in~\autoref{tab:table_steg_ind}, the comparison result demonstrates that VisCode can embed information in various types of chart images with better visual quality.
Besides, we design a user study to evaluate the visual effects of the results by human judgments. We select 15 images from the test dataset as the input to generate steganographic results by 3 methods and recruit 30 users to choose the best one. VisCode achieves a mean proportion of being selected of $83.6\%$, while SteganoGAN achieves $15.1\%$ and StegaStamp achieves $1.3\%$. Our method is rated higher than others.

\begin{figure}[htb]
  \centering
  \subfloat[PSNR]{\includegraphics[width=.333\linewidth]{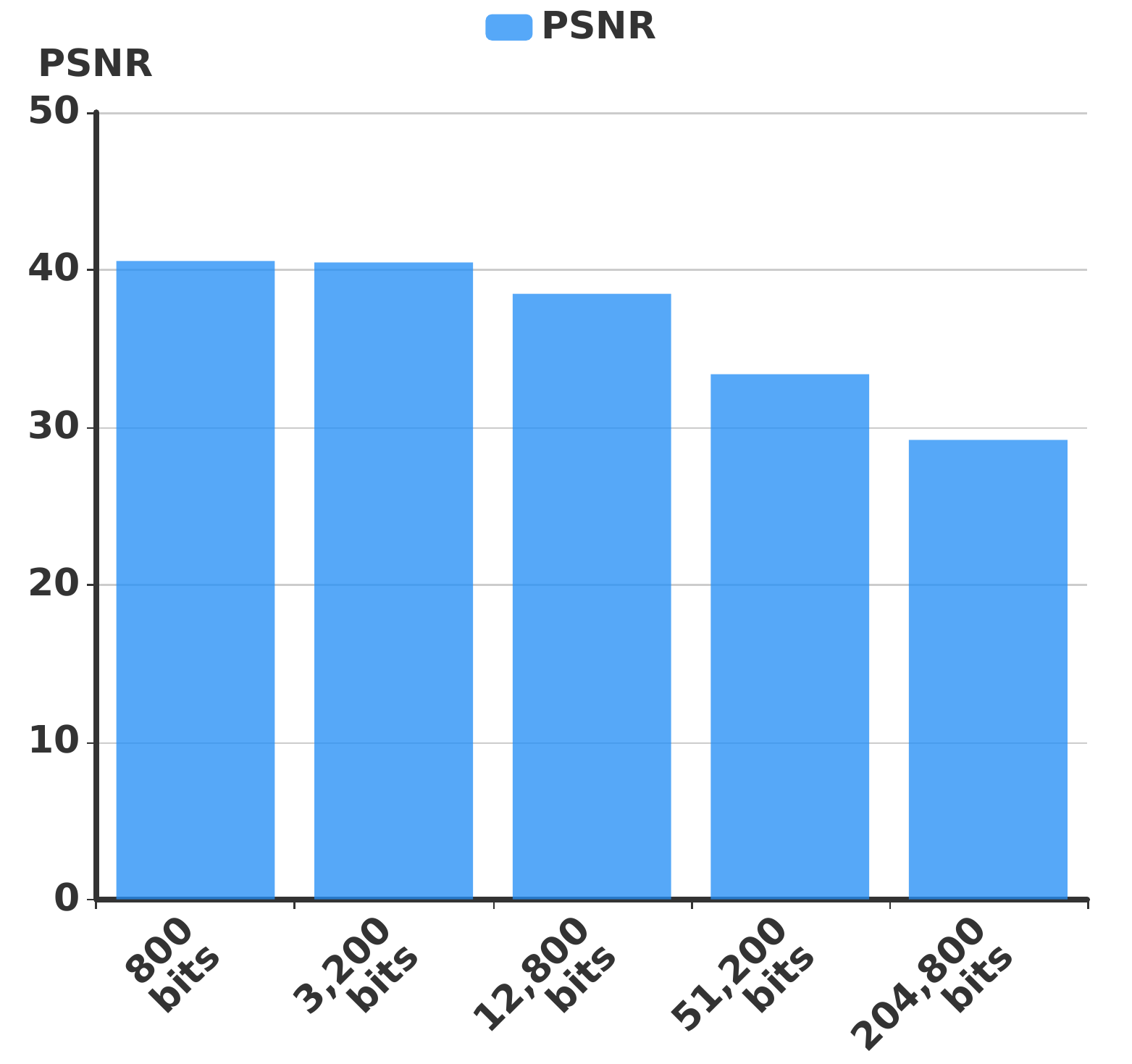}}
  \hfil
  \subfloat[SSIM]{\includegraphics[width=.333\linewidth]{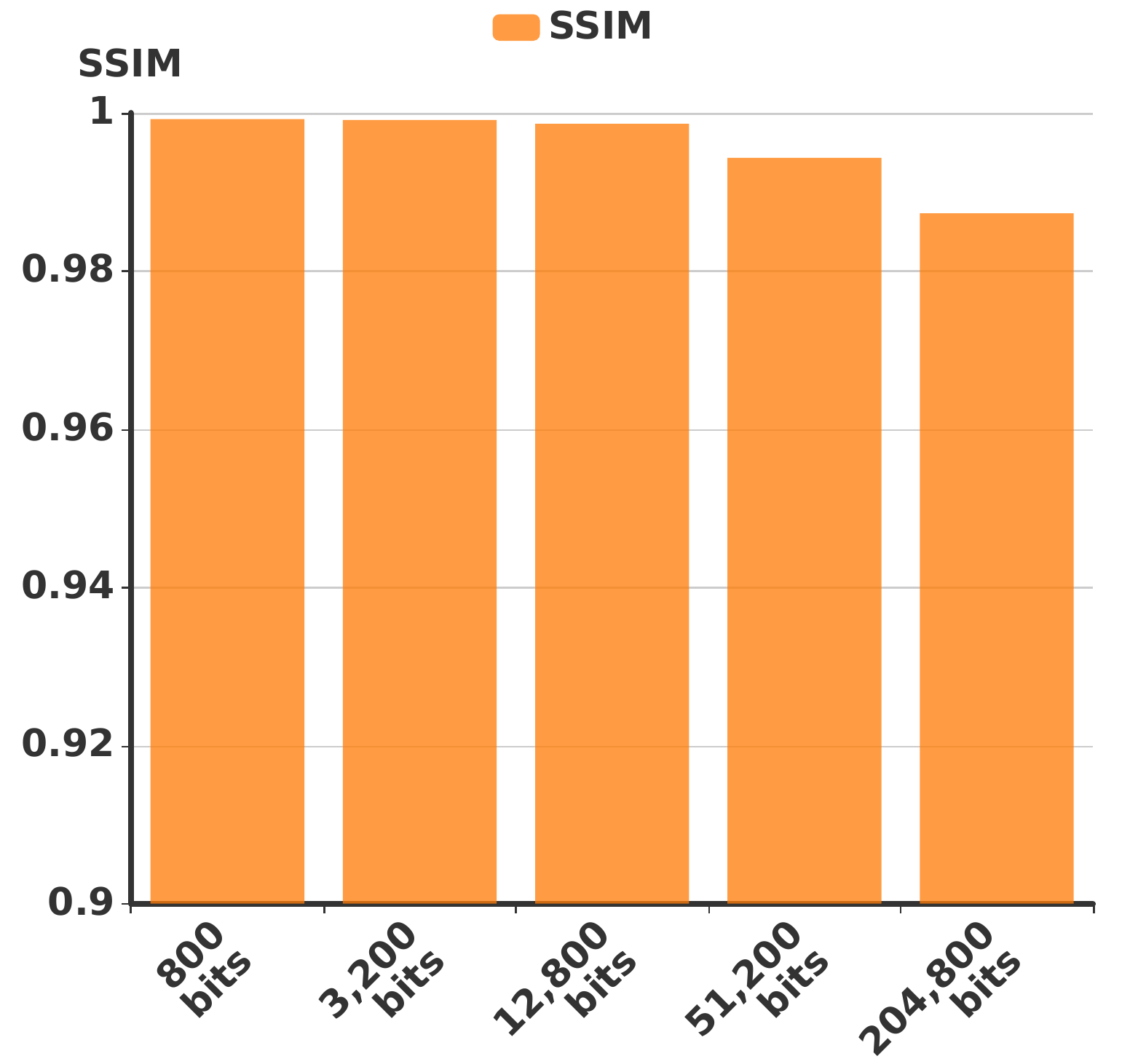}}
  \hfil
  \subfloat[LPIPS]{\includegraphics[width=.333\linewidth]{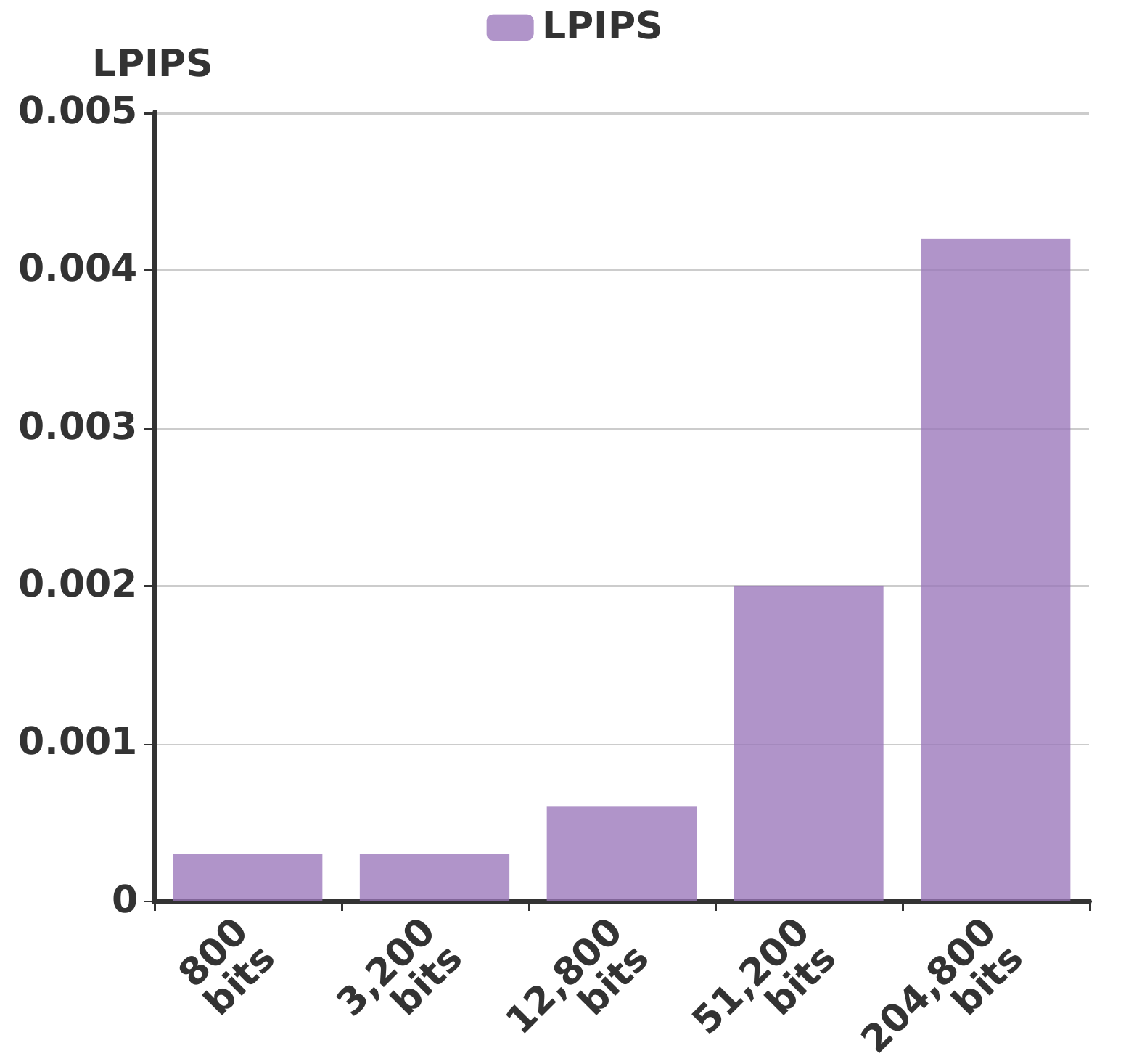}}
  \hfil
  \vspace{-6pt}
  \caption{\label{fig:steg_index}Steganography index evaluation demonstrates that VisCode can encode large-scale data while preserving the perceptual quality of the visualizations.}
\end{figure}

Our VisCode system is designed to store an extensive amount of information. Embedding large data is challenging due to the tradeoff between the visual quality of encoded images and the amount of data.
Therefore, we evaluate the quality of encoded images with different input text lengths. The resolutions of the test chart images are set to $1600 \times 1600$.
~\autoref{fig:steg_index} reports the results of our steganography index evaluation, where X-axis represents the number of bits of input text, and Y-axis shows the average values of associated image quality metrics. Our method achieves 29.18 PSNR, 0.9873 SSIM, and 0.0042 LPIPS on average for embedding $204,800$ bits of information. It demonstrates that VisCode can encode large-scale data while preserving the perceptual quality of the visualizations.


\begin{table}[htp]

\caption{Evaluation results of steganography indices}
\newcolumntype{M}[1]{>{\centering\arraybackslash}m{#1}}
\renewcommand\arraystretch{1.1}
\centering
\small

\begin{tabular}{@{}|M{1.3cm}M{1.0cm}|M{0.5cm}M{0.5cm}M{0.5cm}|M{0.5cm}M{0.5cm}M{0.5cm}|@{}}

\bottomrule
\hline
\multirow{2}{*}{Method}& \multicolumn{1}{|c|}{ \multirow{2}{*}{Size} } &
\multicolumn{3}{c|}{100 bits}&\multicolumn{3}{c|}{3200 bits}\cr\cline{3-8}
&\multicolumn{1}{|c|}{}&$P \uparrow$&$S \uparrow$&$L \downarrow$&$P \uparrow$&$S \uparrow$&$L \downarrow$\cr
\hline
StegaStamp &\multicolumn{1}{|c|}{ \multirow{3}{*}{$400 \times 400$} }&34.09&0.9312&0.0413&-&-&-\cr \cline{1-1} \cline{3-8}
SteganoGAN &\multicolumn{1}{|c|}{}&35.26&0.9386&0.0247&35.19&0.9380&0.0249\cr \cline{1-1} \cline{3-8}

\multirow{2}{*}{VisCode}&\multicolumn{1}{|c|}{}&{40.63}&{0.9959}&{ 0.0007}&{39.19}&{0.9930}&{  0.0008}\cr\cline{2-8}
{}& \multicolumn{1}{|c|}{ {$800 \times 800$} } &{\bf 42.08}&{\bf 0.9990}&{\bf 0.0003}&{\bf 41.70}&{\bf 0.9987}&{\bf 0.0004}\cr
\hline

\toprule

\end{tabular}

\label{tab:table_steg_ind}
\end{table}



\begin{table}[htb]
\caption{Evaluation results of steganography defense.}
\newcolumntype{M}[1]{>{\centering\arraybackslash}m{#1}}
\renewcommand\arraystretch{1.2}
\centering
\small

\begin{tabular}{@{}M{3.3cm}M{3.3cm}|M{1.2cm}M{1.2cm}M{1.2cm}M{1.2cm}|@{}}

\bottomrule
\multicolumn{2}{|c|}{Method} & {Watermark} & {Brightness} & {Rotation} & {JPEG}  \\
\hline

\multicolumn{2}{|c|}{LSB~\cite{mielikainen2006lsb}} & {$0.02\%$} & {$0\%$} & {$0\%$} & {$0\%$} \\
\hline

\multicolumn{2}{|c|}{DeepSteg~\cite{baluja2017hiding}} & {$3.45\%$} & {$32.75\%$} & {$100\%$} & {$9.8\%$} \\
\hline

\multicolumn{2}{|c|}{\textbf{VisCode (Ours)}} & \textbf{55.29$\%$} & \textbf{32.79$\%$} & \textbf{100$\%$} & \textbf{10.2$\%$} \\
\toprule

\end{tabular}

\label{tab:Eva_idx}
\end{table}

\begin{figure}[htb]
  \centering
  \subfloat[Encoding]{\includegraphics[width=.49\linewidth]{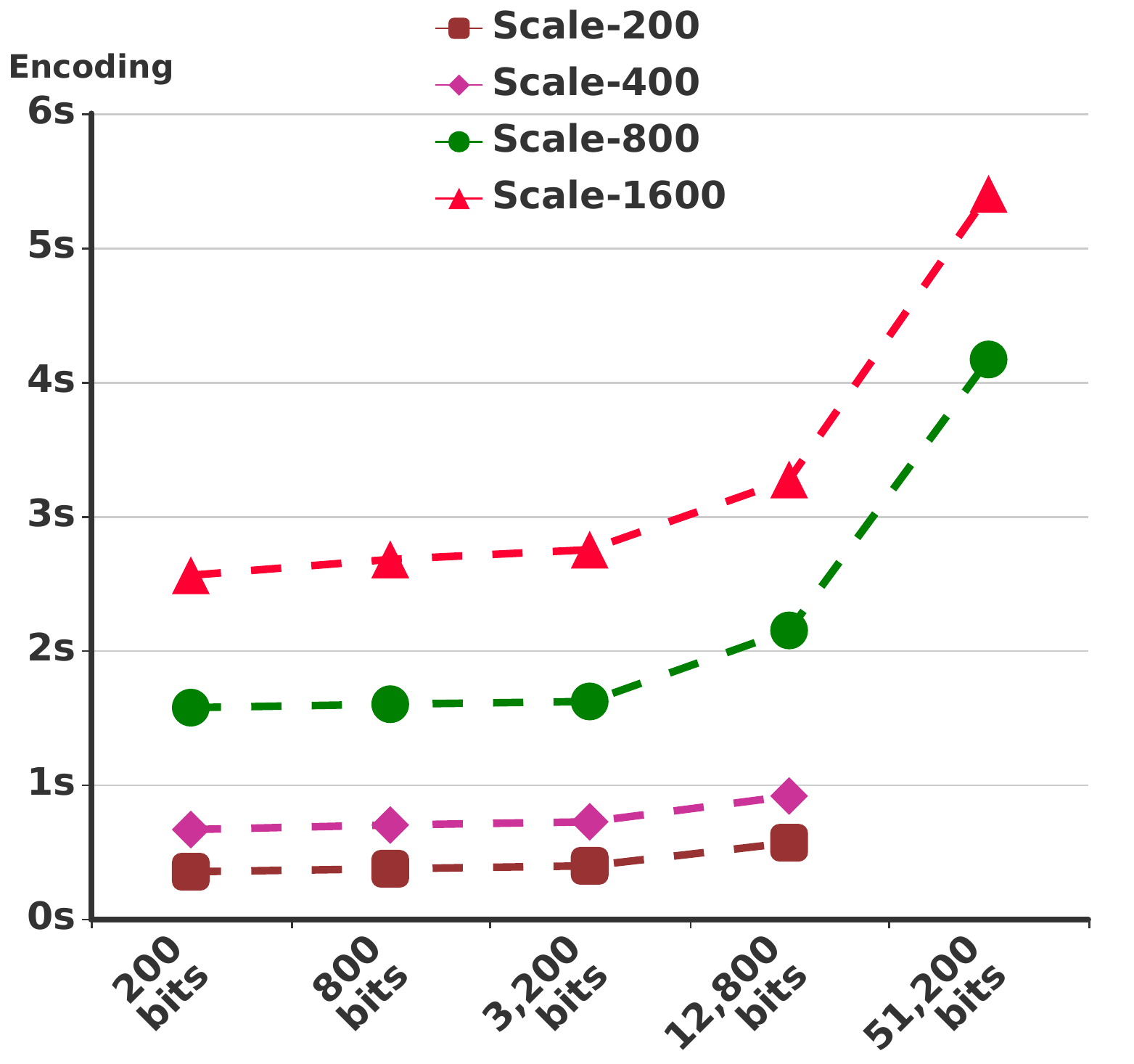}}
  \hfil
  \subfloat[Decoding]{\includegraphics[width=.49\linewidth]{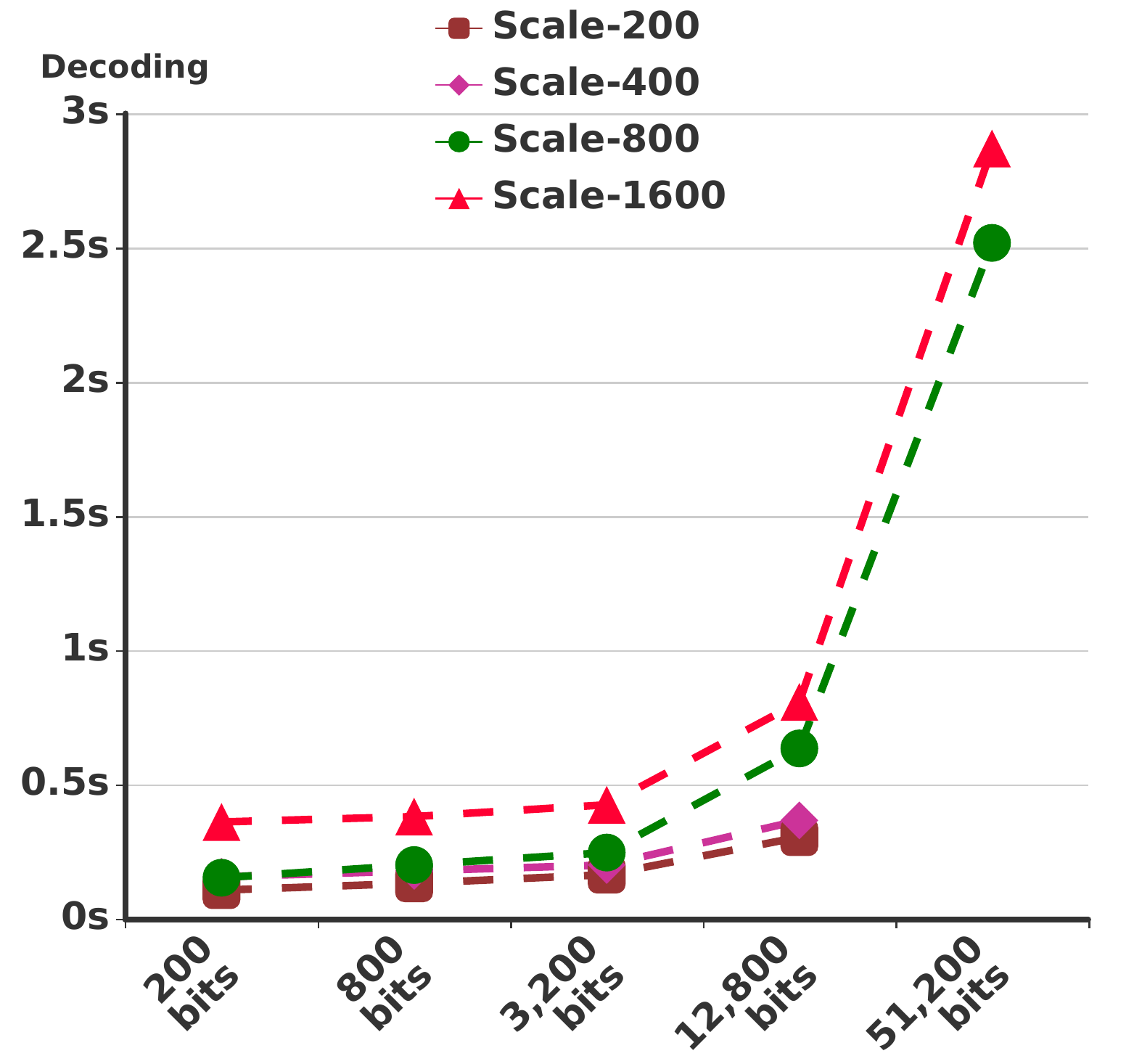}}
  \hfil
  \vspace{-3pt}
  \caption{\label{fig:time_perf}Time performance comparison of encoding and decoding.}
\end{figure}

\begin{figure}[htb]
  \centering
  \includegraphics[width=1.0\columnwidth]{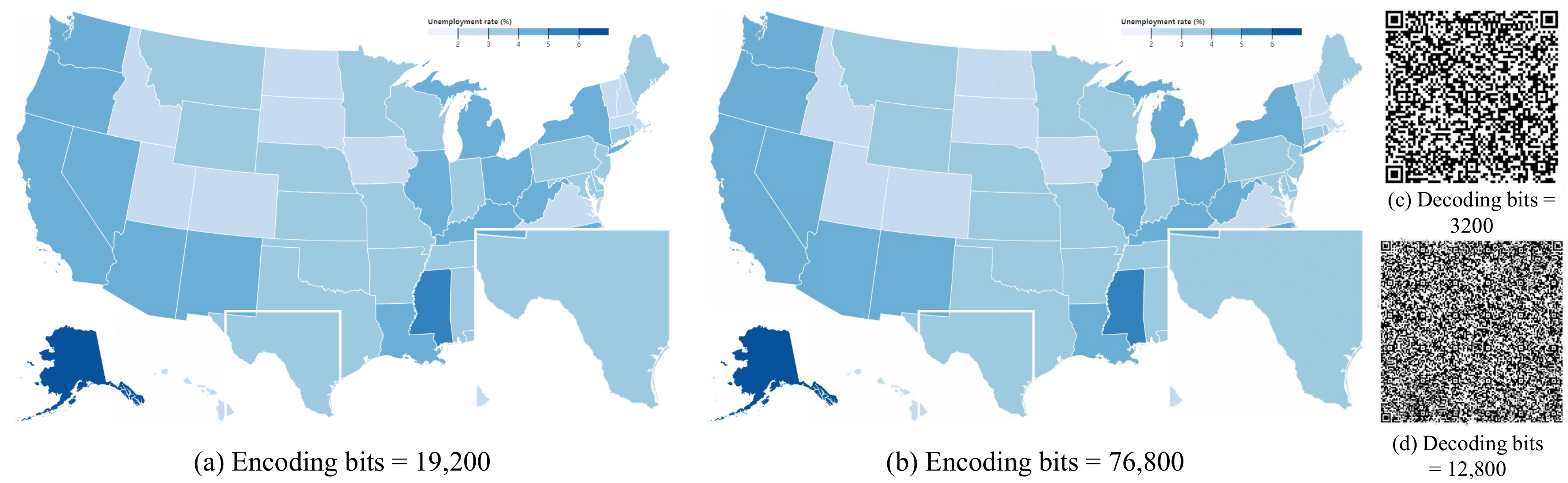}
  \vspace{-12pt}
  \caption{\label{fig:fail_samp}The sample steganographic images generated by VisCode and decoded results with various bits of data. The input chart image is with a resolution of $420 \times 670$. There are fewer visible artifacts in~\autoref{fig:fail_samp}a than that in~\autoref{fig:fail_samp}b. Moreover,~\autoref{fig:fail_samp}c is decoded correctly, while embedding too much information may cause a high decoding error rate, as shown in~\autoref{fig:fail_samp}d.
  }
\end{figure}

\subsection{Steganography Defense}
If we use normal digital transmission without large deformations of the coded images, the decoding text recovery accuracy achieves over $90\%$. When users share encoded chart images through digital transmission, corruption, or attack may occur. Therefore, we analyze four typical scenarios: adding watermark, adjusting brightness, applying rotation and applying JPEG compression. For each image in the test dataset, we first generate the encoded image with text of $12,000$ bits and then apply the following operations: (1) Adding a visible watermark in the same position of the encoded image. (2) Adjusting $ \pm 10\% $ brightness of the image. (3) Rotating the image ${180^\circ }$ clockwise. (4) Applying JPEG compression to the image with quality factor $ Q = 90 $.

We use Text Recovery Accuracy (TRA) to assess the performance of our VisCode model in the decoding stage. TRA is defined as the proportion of recovered characters in all input characters. 
Since it is difficult to embed large-scale data in chart images using StegaStamp~\cite{tancik2020stegastamp} and SteganoGAN~\cite{zhang2019steganogan}, we compare VisCode with other two methods LSB~\cite{mielikainen2006lsb} and DeepSteg~\cite{baluja2017hiding}. In this part, we maintain the diverse resolutions of the testing dataset from $300 \times 300$ to $3000 \times 3000$, as mentioned in ~\autoref{subsec:DV_dataset}. 

~\autoref{tab:Eva_idx} shows the comparison result in steganography defense evaluation. LSB is a traditional steganography method using fixed principles, unable to the defense any perturbations. We find that the performance of VisCode in adjusting brightness, applying rotation and JPEG compression is close to DeepSteg~\cite{baluja2017hiding}, since the three manipulations apply changes on all pixels. Watermarks apply local changes to the image, which may destroy the important features of the image. In this case, VisCode outperforms DeepSteg~\cite{baluja2017hiding} by embedding information in insensitive regions.

\subsection{Time Performance}
The encoder-decoder network is integrated into VisCode with several other processes for interactive applications.
We evaluate time performance for VisCode system of different scales of input chart images and text in~\autoref{fig:time_perf}. X-axis represents the number of bits of input text, and Y-axis shows the average time (in seconds) taken by the complete encoding stage (~\autoref{fig:time_perf}a) and decoding stage(~\autoref{fig:time_perf}b).
Lines of different colors represent different resolutions of the input chart images, ranging from ${\rm{200}} \times {\rm{200}}$ to ${\rm{1600}} \times {\rm{1600}}$. The environment configurations are introduced in~\autoref{sec:apps}.

In the encoding stage, we observe that the time performance is relatively stable for encoding different lengths of information to images with the same resolution. The complete encoding stage includes predicting the visual importance map of the input chart image, localizing the optimal regions, generating QR codes, sending corresponding image pairs to the encoder network, and obtaining the encoded chart image. 
When the resolution of images is constant, the increased time is mainly spent on generating QR codes, while the runtime of the encoder network is very fast. On the other hand, the leading cause of different time performance between images with diverse resolutions is the calculation of the visual importance map and the embedding region proposals. In the decoding stage, the time cost of the different resolution of images is similar owing to the fast decoder network, while larger message needs more time to convert QR codes to text with the error-correction scheme.

\section{Conclusion}
\label{sec:con}

We have proposed an approach for encoding information into a static visualization image. Our applications implemented with VisCode are designed to support metadata embedding, source code embedding, and visualization retargeting. We have designed an encoder-decoder network to effectively encode large amounts of information into visualization images such that the original data can be decoded with high accuracy. The presented applications and evaluations demonstrate that VisCode can offer intuitive and effective information embedding for various types of visualizations.

The current version of VisCode has certain limitations in addressing large deformations of the coded images. In addition, for a fixed visualization image size, there is an upper limit on the amount of information that can be embedded, which reduces the effectiveness of VisCode. If we embed too much information in a chart image, which corresponds to more text in each QR code and lower error correction capability, it may cause a high decoding error rate. With the same image size and encoded data bits, images with more complex texture and rich color content are easier to decode successfully.

In the future, we plan to propose a more flexible framework to ensure a high decoding success rate while adapting to visualization image deformation. Camera capturing and light-aware steganography, as shown in the work of Wengrowski and Dana~\cite{wengrowski2019light}, is a potential research direction that can extend the application domain of VisCode. We also plan to collaborate with designers and developers to obtain feedback in the actual design and manufacturing applications~\cite{Zhou2019survey} to improve the user experience of VisCode.

\acknowledgments{
The authors wish to acknowledge the support from NSFC under Grants (No. 61802128, 61672237, 61532002).
}

\bibliographystyle{abbrv-doi-narrow}

\end{document}